\documentclass[review,1p,times]{elsarticle}

\usepackage{amsmath,amsfonts,amssymb,amsthm}
\usepackage{algorithm2e}
\usepackage{array}
\usepackage{textcomp}
\usepackage{stfloats}
\usepackage{xurl}
\usepackage{verbatim}
\usepackage{graphicx}
\usepackage{balance}
\usepackage{paralist}
\usepackage{booktabs}

\usepackage{tikz}
\usepackage{pgf}
\usepackage{pgfplots}
\pgfplotsset{compat=1.18}
\usepackage{tikzscale}
\usepackage{calc}
\usetikzlibrary{calc}
\usepackage{tikz-dimline}
\usepackage{multirow}
\usepackage{caption}
\usepackage{subcaption}

\usepackage{algpseudocode}
\usepackage{paralist}
\usepackage{listings}
\usepackage[T1]{fontenc}
\usepackage{makecell}
\usepackage[dvipsnames]{xcolor}
\usepackage{color}
\usepackage{ulem}

\definecolor{maroon}{rgb}{0.5,0,0}
\definecolor{darkgreen}{rgb}{0,0.5,0}
\lstdefinelanguage{XML}
{
  basicstyle=\ttfamily,
  morestring=[b]",
  moredelim=[s][\bfseries\color{maroon}]{<}{\ },
  moredelim=[s][\bfseries\color{maroon}]{</}{>},
  moredelim=[l][\bfseries\color{maroon}]{/>},
  moredelim=[l][\bfseries\color{maroon}]{>},
  morecomment=[s]{<?}{?>},
  morecomment=[s]{<!--}{-->},
  commentstyle=\color{darkgreen},
  stringstyle=\color{blue},
  identifierstyle=\color{red},
  tabsize=1
}
\usepackage{textcomp}
\usetikzlibrary{shapes}

\def\BibTeX{{\rm B\kern-.05em{\sc i\kern-.025em b}\kern-.08em
    T\kern-.1667em\lower.7ex\hbox{E}\kern-.125emX}}
\DeclareMathOperator*{\argmin}{arg\,min}

\newcommand{\set}[1]{\ensuremath{\left\{ #1 \right\}}}
\newcommand{\setof}[2]{\ensuremath{\left\{ #1 \middle| #2 \right\}}}

\newcommand{\fnct}[1]{\ensuremath{#1(\cdot)}}

\newcommand{\parconv}[3]{{\ensuremath{\sigma_{{#1}|{#2}}({#3})}}}
\usepackage{tikz}
\usetikzlibrary{automata, positioning, arrows}
\tikzset{
  ->, % makes the edges directed
  >=stealth', % makes the arrow heads bold
  node distance=3cm, % specifies the minimum distance between two nodes. Change if necessary.
  every state/.style={thick, fill=gray!10}, % sets the properties for each 'state' node
  initial text=$ $, % sets the text that appears on the start arrow
}
\makeatletter
\def\Expect{\mathop{\operator@font E}\nolimits}
\makeatother

\newcommand{\magnitude}[1]{\ensuremath{\left\lvert #1 \right\rvert}}

\newcommand{\candlanes}[2]{{\ensuremath{\mathcal{C}(#1,#2)}}}
\newcommand{\neighlanes}[2]{{\ensuremath{\mathcal{N}(#1,#2)}}}

\newcommand{\dpos}[1]{{\ensuremath{\mbox{\textsc{pos}}(#1)}}}

\newcommand{\drones}[2]{{\ensuremath{\Delta(#1,#2)}}}
\newcommand{\dronesunion}[1]{{\ensuremath{\Delta(#1)}}}

\newcommand{\hopdistance}[2]{{\ensuremath{\left[#1,#2\right]_1}}}

\journal{Computer Communications}
\begin{document}
\begin{frontmatter}

\title{\protect PB-STDG: A Prediction-Based Short-Term Decentralized Greedy Guidance Algorithm for a Drone Road System}

\author[labelzqaw]{Zhouyu Qu}
\author[labelzqaw]{Andreas Willig}
\author[labelxw]{Xiaobing Wu}

\affiliation[labelzqaw]{organization={Dept.\ of Computer Science and
    Software Engineering, University of Canterbury},
            addressline={Private Bag 4800}, 
            city={Christchurch},
            postcode={8140}, 
            country={New Zealand}}

\affiliation[labelxw]{organization={Wireless Research Centre, University of Canterbury},
            addressline={Private Bag 4800}, 
            city={Christchurch},
            postcode={8140}, 
            country={New Zealand}}

%===========================================================================
%                 Abstract
%===========================================================================   
\begin{abstract}
In recent years, Unmanned Aerial Vehicles (UAVs) or drones have been increasingly adopted in urban environments for applications such as parcel delivery, infrastructure inspection, emergency response, and drone light shows. A non-negligible issue is how to manage the increasing number of drones operated by different entities, to enable them to cooperatively avoid potential collisions and determine conflict-free short-term flight paths in urban airspace. This paper presents a Prediction-Based Short-Term Decentralized Greedy (PB-STDG) guidance algorithm for a structured Drone Road System (DRS). PB-STDG extends the original STDG algorithm by introducing a prediction mechanism that enables drones to anticipate the decisions of neighboring drones using additional information shared in beacon packets, aiming to address the over-conservative behavior observed in the STDG algorithm. Several simulation scenarios are used to evaluate and compare the proposed algorithm with STDG. The results show that PB-STDG improves traffic efficiency while maintaining a safety level comparable to that of STDG.
\end{abstract}


% \begin{highlights}
% \end{highlights}


\end{frontmatter}


\bibliographystyle{abbrvnat}
\setcitestyle{numbers,open={[},close={]}}

% ==========================================================================================
%                        Intro
% ==========================================================================================
\section{Introduction}

Continuous advances in drones, or unmanned aerial vehicle (UAV), technology have significantly expanded the range of drone applications, attracting growing interest from both academia and industry. Although drones were initially developed for military operations \cite{ko2021drone, bousquet2018eye}, their applications have gradually expanded into a wide range of civilian and commercial fields, such as aerial photography \cite{jung2022improved,ri2024drone,xiao2022aerial}, emergency response \cite{soni2024medical}, and infrastructure inspection \cite{lekidis2022electricity, raja2022adversarial}. More recently, drone applications have expanded into delivery and entertainment. Commercial drone delivery services are now being used for last-mile parcel distribution by companies such as Amazon Prime Air \cite{amazon_prime_air} and Google Wing \cite{wing_delivery}. Food delivery by drones has also entered commercial operation in cities including Shenzhen, China, where Meituan operates urban drone delivery routes \cite{meituan_drone}. Beyond delivery, large-scale drone light shows have become a popular feature of public celebrations, such as the drone show program at Darling Harbour in Sydney \cite{darling_harbour_drone}, where coordinated drone swarms are deployed for regular night-time performances.

With the growth of these drone applications, the number of deployed drones is continuously increasing \cite{dii2025}. These drones generally share the same low-altitude airspace, particularly in urban environments. Unlike traditional ground traffic systems, most current drone operations are conducted in an unstructured and free-flight manner. Each drone independently plans its trajectory based on its own algorithms and available information. Although such an approach provides flexibility, it becomes increasingly difficult to maintain safety and manage large-scale drone operations involving different operators and service providers. Conflicts between competing flight paths may lead to traffic congestion or even collisions as the number of drones increases, which can degrade overall system efficiency and pose risks to public safety.

To address this challenge, several studies on structured airspace management have been proposed, aiming to organize low-altitude airspace into predefined aerial structures and constrain drone movements within them. A representative example is the NASA drone traffic management concept \cite{jang2017concepts}. In this framework, the low-altitude airspace is organized into multiple layers based on altitude. High-speed fixed-wing UAVs are typically assigned to higher layers, while low-speed rotary-wing UAVs operate in lower layers. Within each layer, the airspace is further structured into different types of aerial structures, including sky-corridors, sky-tubes, and sky-lanes. Drones are allowed to fly freely within sky-corridors, whereas sky-tubes impose directional constraints on movement. Sky-lanes provide an additional level of restriction by regulating lateral motion, preventing drones from cruising in parallel within the same lane.

Several studies have explored layer-based approaches for drone traffic management, using different altitude layers to regulate drone movement and heading directions \cite{aerospace8020038,svaigen2022biomixd,9256627}. Other studies have proposed more structured drone road systems inspired by ground transportation networks, including corridor-based and lane-level structures \cite{ijgi10050338,9476745}. Building on these concepts, \cite{qu2025drone} proposes a Drone Road System framework consisting of structured roads and ramps. All road segments are designed to be unidirectional, with a defined number of lanes and an allowed speed range, to support safe and predictable drone movement. The system can also be encoded in XML format, enabling flexible updates and efficient distribution across different platforms.

While drone traffic management systems such as drone road systems provide a structured framework for organizing low-altitude airspace, conflicts between drones can still occur within the constrained network, particularly under high-density traffic conditions. In such environments, drones may compete for optimal routes, and different types of conflicts may arise due to speed differences or directional changes during navigation. Therefore, effective drone guidance and collision avoidance mechanisms remain essential.

A large body of research has been devoted to UAV navigation and collision avoidance in free-flight environments. These include approaches such as \cite{farid2022modified} and \cite{wang2022trajectory}, which determine short-term collision-free paths for UAVs in environments containing obstacles and other dynamic constraints. The Particle Swarm Optimization (PSO) approach \cite{abhishek2020hybrid} iteratively updates drone positions based on individual and global best solutions. Another widely studied approach is Deep Reinforcement Learning (DRL) \cite{han2025deep, almahamid2025agile, liu2025flexible}, which learns drone control policies through interaction with the environment. These methods typically use sensory inputs such as images, video streams, or onboard sensor data to generate control actions via deep neural networks. By optimizing a reward function, drones can learn collision-free navigation in dynamic and complex environments. 

Specifically, several studies have investigated safe guidance and collision avoidance algorithms for drone delivery scenarios. In \cite{droneincity3}, three collision avoidance strategies are proposed to manage large numbers of drones delivering goods to random destinations in urban environments. Similarly, \cite{droneincity1} presents a centralized navigation approach for a fleet of drones to achieve safe point-to-point parcel delivery missions. In addition, \cite{droneincity2} introduces two centralized algorithms to ensure safe delivery operations within a simplified two-layer airspace structure.

However, these approaches suffer from several limitations: (i) they often rely on global information of other agents, (ii) they require multiple types of onboard sensors, and (iii) they are primarily designed for free-flight environments and cannot be directly adapted to structured airspace systems. In the large-scale drone traffic system we focus on, where drones are owned by different operators and perform diverse tasks, these limitations become more critical. Specifically, global coordination is difficult to achieve in practice, as it is unrealistic for each drone to obtain or maintain global knowledge of all other drones in the city-scale airspace. Moreover, different navigation algorithms across different drones further increase system uncertainty, since individual drones are unable to accurately predict the behavior of others under inconsistent or unknown algorithms. They will also make different assumptions about the available sensors, and in this paper, we make very minimal assumptions, requiring only GPS and wireless communication.

Based on these challenges and assumptions, a research gap can be identified in the guidance of large-scale drone traffic under limited coordination. Existing approaches often rely on assumptions about the available sensing, communication, or coordination capabilities of drones. However, in a large-scale drone road system, drones may be operated by different manufacturers or service providers and may therefore have limited ability to coordinate with one another. There is therefore a need for a guidance approach that can manage interactions among a large number of independently operating drones without relying on extensive coordination or additional sensing capabilities.

To address this research gap, a baseline guidance algorithm for structured drone road systems is required that can be applied to different types of drones while relying on minimal information and limited sensing capabilities. In our previous studies \cite{drones6120411, qu2026stdg}, STDG was proposed to address drone route navigation and collision avoidance challenges on the standardized Drone Road System introduced in \cite{qu2025drone}. The STDG algorithm relies only on position and velocity information exchanged between drones through beacon packets, and provides short-term, coordination-free, and decentralized guidance. Drones periodically execute STDG to make short-term lane and speed decisions based on the current traffic conditions and potential collision risks.

Previous simulation results for STDG revealed a key limitation: significant speed loss when avoiding potential collisions. Since the algorithm is coordination-free (i.e., drones only exchange beacon packets, with no other form of coordination) and only position and velocity information is shared, it is difficult for drones to accurately estimate or predict the decisions of others. As a result, lane switching may be suppressed when multiple neighboring drones simultaneously select the same lane to switch. In such cases, drones tend to remain in their current lanes and may further reduce their speed or even come to a complete stop until the surrounding traffic conditions change. This behavior significantly degrades traffic throughput and system efficiency. In our previous study, the algorithm settings that allow some potential risks to be disregarded result in an increase in average speed from around 7 m/s to 10 m/s, approaching the preferred speed of 12.5 m/s, with only a slight increase in the collision rate (i.e., from 1 to 5 collisions among 150,000 drones). The results show that STDG has room to improve its average speed, as a substantial speed increase can be achieved by slightly relaxing its treatment of potential collision risks with only a small increase in actual collisions. 

This motivates the prediction of neighboring drones' future actions to better distinguish potential risks from likely collisions, thus allowing drones to make more effective lane-switching and speed-selection decisions without unnecessary stops or slowdowns and avoiding additional collisions that may occur in such situations. 

In this paper, we focus on designing a prediction-based extension to the previous STDG guidance algorithm. The proposed approach extends the beacon structure by including additional safety information, which allows drones to predict the movement decisions of their neighbors over the next short period. This capability helps reduce situations where drones tend to remain in their current lanes and encourages necessary lane changes, either to overtake slower predecessors or to move towards their target lanes without prolonged stopping.

Extensive simulations are conducted using the same simulation framework and largely the same experimental configuration as the original STDG study. The results show that, under the same traffic scenario, the proposed prediction-based algorithm substantially reduces the speed loss while maintaining a close-to-zero collision rate. Additional simulations are performed under different road-system configurations and algorithm parameter settings to comprehensively evaluate PB-STDG against the original STDG algorithm.

The main contributions of this paper are summarized as follows.
\begin{itemize}
\item We propose PB-STDG, a prediction-based extension of the original STDG algorithm that enables drones to predict the movement decisions of neighboring drones. The proposed approach reduces the tendency of multiple drones to remain in their current lanes during concurrent lane switching and, in the tested complex DRS, increases the average speed from 10.28 m/s to 12.18 m/s while maintaining a close-to-zero collision rate.

\item We design an extended beacon structure and an event-driven beacon transmission mechanism that provide the guidance algorithm with additional safety information for predicting the decisions of neighboring drones, with a modest increase in beacon packet size.

\item We conduct extensive simulations under different drone densities and algorithm configurations to evaluate the proposed algorithm and compare it with the STDG algorithm. The results demonstrate the effectiveness of the proposed approach across different traffic conditions and algorithm settings.
\end{itemize}

The remainder of this paper is organized as follows. Section \ref{sec:background} briefly introduces the DRS adopted in this work and reviews the baseline STDG algorithm on which PB-STDG is built. Section \ref{sec:system model} describes the system settings and assumptions, including the wireless communication and networking settings, beacon content, and beacon transmission mechanism. Section \ref{sec:algorithm} presents PB-STDG in detail. Section \ref{sec:results} describes the simulation assumptions and configurations and evaluates PB-STDG against STDG through extensive simulation studies. Finally, Section \ref{sec:conclusion} concludes the paper and discusses future research directions.

\section{Background}
\label{sec:background}

The prediction-based algorithm proposed in this paper is built upon the drone road system (DRS) \cite{qu2025drone} and the Short-Term Decentralized Greedy (STDG) algorithm \cite{qu2026stdg} developed in previous work.  

\subsection{Drone Road System}
\label{sec:drs}

For large-scale drone traffic management, the drone road system introduces a common framework to construct predefined routes. This framework simplifies traffic management and collision avoidance by making the movements of individual drones more predictable. The predefined road system can be represented in an XML file, which simplifies both road network updates and the transmission of updated information to drones currently cruising within the system.

\begin{figure}[htbp]
  \centering
  \includegraphics[scale=0.3]{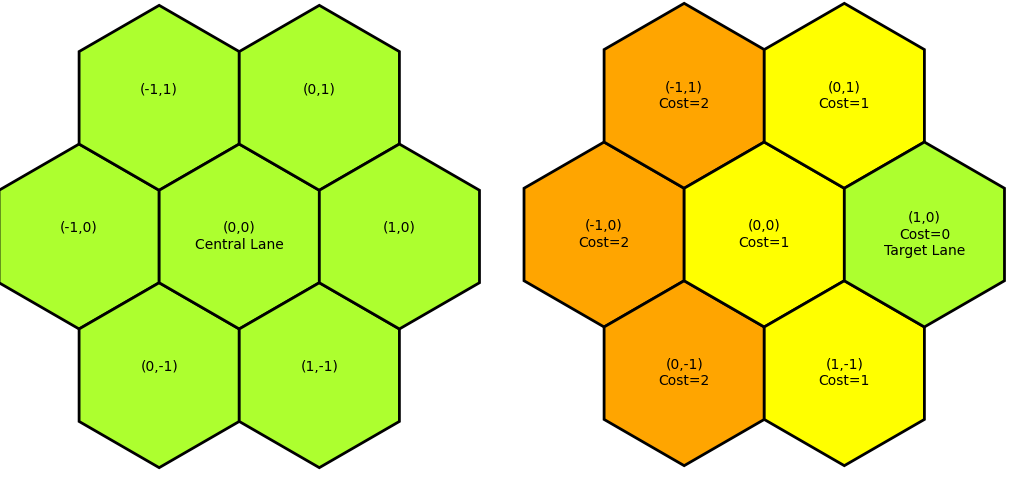}
  \caption{Left: Simple lane arrangement example, identifier (0,0) is the central lane of the road. Right: Example of position-cost calculation.}
  \label{fig:lane arrange}
\end{figure}

Each drone road system consists of one or multiple roads and zero or multiple ramps. Both roads and ramps are unidirectional, meaning that drones within the same road or ramp follow the same travel direction. Road is the primary component of the system and consists of one central lane and multiple parallel lanes arranged in a hexagonal grid topology. Each lane is assigned a unique identifier $(i,j)$ and can have up to six neighboring lanes (shown in Figure \ref{fig:lane arrange}), with the identifiers $(i+1,j)$, $(i-1,j)$, $(i,j+1)$, $(i,j-1)$, $(i+1,j-1)$, and $(i-1,j+1)$. Since roads may be curved, an arc-length parameter $s$ is introduced to represent the relative position of a drone along a lane. The value of $s$ represents the distance traveled by the drone from the beginning of the lane along the lane.

Ramp is a special type of road that consists of a single lane and is used either to connect two roads or to provide additional entry and exit access to the drone road system. When constructing a ramp, one or two connection points are defined on existing road lanes. If only one connection point is specified, the ramp is referred to as an on/off-ramp, where the other end serves as either an entry or exit of the system. Otherwise, the ramp connects two lanes and is referred to as a connecting ramp.

During its lifetime in the system, a drone first enters the network through either a road entry point or an on-ramp, and its end-to-end path planning is done in advance. In this study, we assume that the plan is never altered afterwards. It then navigates through the system by moving along lanes and may perform lane switching to overtake slower drones or to reach connected lanes leading to other roads or ramps. In this process, a drone can only directly switch to one of its neighboring lanes and then repeat the lane-switching maneuver until it reaches its target lane. Finally, the drone reaches its destination road or off-ramp and exits the system at the corresponding endpoint.

\subsection{Short-Term Decentralized Greedy Algorithm}

The Short-Term Decentralized Greedy (STDG) algorithm serves as a baseline guidance method for the drone road system introduced in Section \ref{sec:drs}. STDG ensures that drones can successfully complete their trips within the road system while providing essential functionalities such as collision avoidance, lane-switching, and speed-selection. The algorithm generates short-term decisions based on the current local traffic situation over a limited time horizon. This enables drones to overtake slower neighbors and progress toward target connection points without missing them.

The inputs of STDG consist primarily of the position and velocity of the current drone and those of neighboring drones, obtained through periodic beacon transmissions. The algorithm is fully decentralized and does not require any explicit coordination beyond beacon exchange. At the start of the algorithm, a drone considers its current lane and all neighboring lanes as candidate lanes, and evaluates a weighted cost function for each candidate. The lane with the minimum cost is selected as the output lane. If the selected lane differs from the current lane, the drone performs a lane-switching action; otherwise, it remains in its current lane. After lane selection, the drone updates its speed based on either its preferred speed or the speed of its immediate predecessor in the selected lane, whichever is lower.

The total cost consists of three components: position cost, speed cost, and collision cost. The position cost is designed to guide drones toward their target connection points and support lane switching decisions toward subsequent roads or ramps. The basic position cost is defined based on the hop distance between the current lane and the target lane associated with the next connection point, and the distance to that connection point. For example, consider a drone located at lane $(0,0)$ with a target lane $(1,0)$, assuming that all neighboring lanes are available and accessible. The basic position cost for each candidate lane is defined as (example shown in Figure \ref{fig:lane arrange}):
\begin{itemize}
    \item \textbf{0}, for lane $(1,0)$
    \item \textbf{1}, for lanes $(0,0)$, $(0,1)$, and $(1,-1)$
    \item \textbf{2}, for lanes $(-1,0)$, $(0,-1)$, and $(-1,1)$
\end{itemize}
If a drone does not have a target connection point (i.e., its destination is the current road/ramp.), its basic position cost is set to zero. The final position cost is further scaled by a multiplicative factor based on the distance between the drone's current position and its target connection point. This factor is  a monotonically non-negative decreasing function of distance, which reduces the influence of position cost for drones that are far from their targets, while increasing the switching priority for drones that are closer to their target connection points.

The speed cost imposes a penalty on drones when their deployable speed is lower than their preferred speed. The deployable speed for each candidate lane is determined by the drone's preferred speed, the speed of its predecessor, and the distance-dependent constraints. Here, the predecessor refers to the nearest ahead drone in the same lane. If the distance to the predecessor is below a predefined threshold and the predecessor's speed is lower than the drone's preferred speed, the deployable speed is set to the predecessor's speed. Otherwise, it is set to the drone's preferred speed. The speed cost is then computed using an increasing function of the gap between the preferred speed and the deployable speed.

The collision cost is designed to prevent lane switching maneuvers that may lead to collisions. Its calculation consists of two steps. First, if any drone is detected in the candidate lane within a predefined distance of the current drone, the collision cost is set to the maximum value. This indicates that the candidate lane is infeasible for switching due to the high risk of an immediate collision. Second, if the cost is not set to the maximum value, the algorithm counts the number of drones in the lanes adjacent to the candidate lane that are within the same predefined distance of the current drone. The collision cost is then set equal to this count. This value represents the potential collision risk caused by multiple drones simultaneously attempting to switch into the same lane. A special rule is introduced for leading drones to break deadlocks that may occur when multiple drones are closely clustered and block one another. A leading drone is defined as the foremost drone that has no other drone ahead of it within a predefined safety distance. Such a drone is assigned the highest lane switching priority, and its cost in the second stage of the cost calculation is always set to zero. This allows the leading drone to switch lanes first, thereby creating opportunities for the following drones to resolve the blockage.

The final cost is calculated as the weighted sum of the three cost components. The corresponding weights are configurable algorithm parameters that can be adjusted for different traffic scenarios.

After selecting the output lane and speed, the drone determines whether it is sufficiently close to its target connection point or the road closing point. If so, and the selected output lane is the current lane, the drone enters a stop state. This decision aims to help the drone avoid missing its target connection point or entering the closed section of the road. In this case, the interval before the next algorithm execution is reduced to allow the drone to leave the stop state more quickly when an opportunity becomes available. Otherwise, the interval is set to its normal value. The drone then maintains the selected speed and stays at, or switches to, the selected output lane until the next execution of the algorithm.

% ==========================================================================================
%                        System Model
% ==========================================================================================

\section{System Model}
\label{sec:system model}

In this section, the scope of this study and the underlying assumptions are described. The system model is adopted from the previous STDG study \cite{qu2026stdg} and is presented here for completeness before introducing the proposed prediction-based algorithm.

% ==========================================================================================
\subsection{Beacon Packet}
\label{sec:system model:beacon}
PB-STDG relies only on information exchanged through beacon packets. In STDG, the beacon mainly includes position and velocity information (several 3-D coordinates) of nearby drones. To better predict the decisions of other drones, the prediction-based algorithm extends the information included in each beacon while maintaining the same periodic beaconing rate. The newly introduced information types are:

\begin{itemize}
    \item \textbf{Preferred speed}: Each drone is assigned a preferred speed representing its optimal cruising speed for maximum energy efficiency based on its weight and physical design. This double value is included in the beacon to allow other drones to estimate the sender's speed cost.
    \item \textbf{Lane being switched to}: If the drone is currently switching lanes, it informs other drones of the lane it is switching to. Otherwise, this field remains empty.
    \item \textbf{Time for next algorithm execution}: This information allows other drones to estimate the likelihood of the sender changing its movement decision in the near future. If a sufficient time gap exists before the next execution, other drones may assume that the sender will maintain its current movement state and will not participate in lane switching competition.
    \item \textbf{Target lane}: If the destination of a drone is located on a different road or ramp from its current position, the drone must switch to the lane containing the connection point leading toward that destination. Sharing this information enables other drones to estimate the sender's position cost.
\end{itemize}

\begin{figure}[htbp]
  \centering
  \includegraphics[scale=0.4]{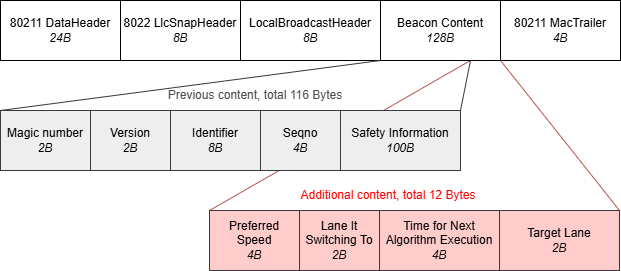}
  \caption{Beacon Content}
  \label{fig:beacon content}
\end{figure}

Figure \ref{fig:beacon content} shows the modified beacon structure and the corresponding length of each field. The original beacon size was 160 bytes, while the proposed extension increases it to 172 bytes. For the main content block, the original size was 116 bytes, and the newly introduced information adds only 12 bytes, extending the size to 128 bytes.

An important assumption of this study is that each drone can accurately determine its current position and velocity using its onboard GPS without measurement uncertainty, and that all drones are time-synchronized via GPS. Assuming perfect GPS measurements allows any remaining information uncertainty to be attributed to the communication process. Furthermore, the drone is assumed to be able to convert the GPS measurements, including latitude, longitude, altitude, speed, and heading, into the corresponding position and velocity representation used in the drone road system (DRS). In the DRS, position information consists of the road identifier, lane identifier, and the arc-length parameter along the current road segment. Velocity refers to the moving speed of the drone along the current road. 
% ==========================================================================================
\subsection{Wireless Communication}
\label{sec:system model:network}

In the STDG system model, drones periodically broadcast beacon packets and obtain information about neighboring drones by receiving their beacons. A parameter, beaconing rate ($br$), is used to represent the number of beacon transmissions per second (Hz). Then, the average beacon interval is approximately $1/br$, although a small amount of randomness is introduced into each interval to avoid synchronization effects.

In the prediction-based algorithm, a key modification to the beaconing mechanism is that a drone transmits an additional beacon immediately after completing an execution of the algorithm, in addition to the regularly scheduled beacon transmissions. This allows other drones to promptly obtain updated information about its movement state after a decision is made, providing more up-to-date information for predicting its subsequent actions.

It is assumed that when an additional beacon is transmitted, the regular beacon timer is reset. Consequently, the next periodic beacon transmission will occur after another interval of approximately $1/br$ seconds. Furthermore, beacon transmissions are assumed to be local broadcasts that do not require acknowledgments or routing protocols. Signal interference in the system is controlled by the beaconing rate and transmit power, and occurs only among beacon transmissions. Excluding external interference allows communication-related uncertainty to be attributed primarily to beacon collisions, packet loss, and transmission delays. Such interference may result in beacon packet loss or transmission delay. This will be the main source of information uncertainty in the simulation. 

The beacon transmission mechanism follows the beaconing approach commonly adopted in IEEE 802.11p-based vehicular ad hoc networks (VANET) systems \cite{karagiannis2011vehicular}. However, the physical-layer parameters are adapted for the drone communication scenario. In particular, a center frequency of 2.4 GHz, a bandwidth of 10 MHz, and a transmission bitrate of 12 Mbps are employed in the simulation. At the MAC layer, the IEEE 802.11 EDCA mechanism is adopted as the default channel access scheme for vehicular ad hoc networks. In this work, only the best-effort access category (AC\_BE) is used. The contention window is configured with a minimum value of 31 and a maximum value of 1023, and the AIFS is set to 7. For the wireless channel model, a log-distance path loss model with a specified path loss exponent is adopted \cite{theodore2002wireless}. External interference, multipath fading, and shadowing effects from buildings or other structures are not considered in this work.

%==========================================================================================
% ==========================================================================================
%                        Guidance Algorithm
% ==========================================================================================
\section{Proposed PB-STDG Algorithm}
\label{sec:algorithm}

The proposed algorithm is a variant of the previous short-term decentralized greedy (STDG) algorithm \cite{qu2026stdg}, which relies on additional shared information in the beaconing packet to reduce the upcoming decision uncertainty of neighboring drones. This algorithm requires drones to operate within the drone road system formalism proposed in \cite{qu2025drone}.

% ==========================================================================================

\subsection{Preliminaries}
\label{sec:algorithm:preliminaries}

%In the existing STDG algorithm, drones make decisions based on the position and velocity information shared in the beaconing packet. When executing the algorithm, drones first select several candidate lanes from all neighboring lanes, and then calculate the speed, position, and collision cost for each candidate. The final output lane is chosen from the candidates with the lowest weighted total cost. By selecting different values of the algorithm parameters, the decision can be either conservative, avoiding most potential risks at the cost of decreased average speed, or aggressive, achieving higher speed and throughput but with an increased possibility of collision. 

One challenge of STDG is the uncertainty of neighboring drones. With limited shared information, it is hard to predict the upcoming actions of neighbors. Thus, to avoid most potential risks, drones should choose conservative decisions, such as slowing down or keeping the current lane, which lowers the average speed and throughput. In addition, under these actions drones tend to stay in outer lanes which have fewer neighboring lanes. Based on the preliminary study (tested DRS is shown in Figure \ref{fig:drsInSim}), the outer lane usage ($\approx 22\%$) is noticeably higher than the inner lane usage ($\approx 13\%$). This leads to low utilization of inner lanes and reduced efficiency of the DRS. 

Based on the observations, achieving the lowest collision rate (<0.01\%) results in an average drone speed of only 7.06 m/s, well below the preferred cruising speed of 12.5 m/s. When the algorithm is configured to disregard most potential risks from neighboring drones, the average speed increases to 11.6 m/s, while the collision rate increases only slightly to 0.36\%. When all potential risks are disregarded, the average speed further increases to 12.3 m/s, with a collision rate of 1.27\%. These results indicate that maintaining a close-to-zero collision rate requires drones to substantially reduce their speed to account for potential collisions that may not actually occur, such as when neighboring drones continue along their current trajectories.

This is the motivation of the proposed prediction-based guidance algorithm. It aims to make safer and more efficient decisions that improve average speed, throughput, and DRS utilization while maintaining the close-to-zero collision rate achieved by the coordination-free STDG approach. We assume that all drones run the same guidance algorithm and know their end-to-end paths, which can be planned by other algorithms, and are also familiar with the entire DRS within which they operate.

\begin{figure}[ht!]
  \centering
  \includegraphics[scale=0.4]{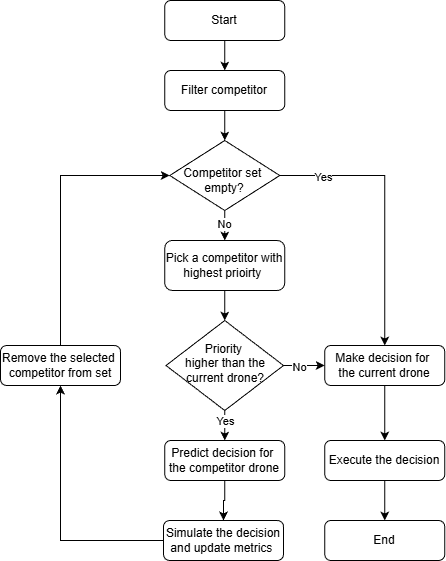}
  \caption{Workflow of PB-STDG}
  \label{fig:workflow}
\end{figure}

At the highest level, the algorithm first filters competitor drones from all nearby drones and predicts a decision for each according to their priority which is mainly determined by its current position. The drones are considered competitors when they would run the algorithm on their side within a short time. After each predicted decision, the algorithm assumes the relevant competitor will make the same decision and then simulates its effect on the current environment and updates relevant parameters for the prediction. This process is repeated until no competitor has higher priority than the current drone. The algorithm then makes the actual decision for the current drone based on iteratively updated parameters for the current drone, specifying the speed and lane to be used until the next invocation. The overall workflow is shown in Figure \ref{fig:workflow}.

% ==========================================================================================

\subsection{The Motivation of Splitting Blocks}
\label{sec:algorithm:motivation}

%One challenge of coordination-free algorithms is inner lane vanishing, which results in lower utilization of some lanes compared to other lanes. From previous simulation results\footnote{(More quantitative data shown in Section \ref{sec:results:position})}, the inner lane usage in a high drone density scenario for a given drone road system (shown in Figure \ref{fig:drsInSim}) is around 13\%, which is lower than the expected value 20\%. Since only the position and velocity information of neighboring drones are available, their future actions cannot be predicted. To avoid potential collisions, drones tend to adopt a conservative strategy, namely maintaining their current lane, which can lead to this phenomenon.

In prediction-based scenarios, drones are assumed to exchange additional information, which improves the accuracy of predicting other drones’ actions. By knowing how others make their decisions, it becomes easier to plan a safe and efficient path for the current drone. However, due to various limitations such as communication range and packet loss, drones cannot obtain complete global information about all drones in the system. Moreover, even neighboring drones may possess different information. In such cases, predictions may not match the actual actions when the predictor’s knowledge differs from that of the predicted drones.

\begin{figure}[htbp]
  \centering
  \includegraphics[scale=0.8]{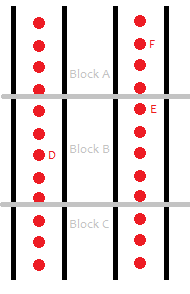}
  \caption{Road division into equal-length blocks for more accurate prediction}
  \label{fig:inner splitted}
\end{figure}

To improve prediction accuracy, we divide each road into multiple equal-length blocks whose boundaries are normal to the central lane. With a known block length $\epsilon_b$ and the distance traveled along the road $L_t$, a drone can determine the block to which it currently belongs as $\left\lfloor L_t/\epsilon_b\right\rfloor$. Drones only predict the actions of others within the same block, while treating drones in other blocks as potential risks. The process is shown in Figure \ref{fig:inner splitted}. Drones D and E belong to the same block, and E has higher priority since it is ahead. When D runs the algorithm, it first simulates E and then makes its own decision based on the simulation result. When E runs the algorithm, it directly makes its own decision due to its highest priority without considering D. During the decision-making and prediction process, drones in other blocks, such as F, are considered unpredictable, and D and E use the previous coordination-free approach to account for them.

In the above process, drone D knows how E will make its decision and already has all relevant information, so this prediction has a higher probability of being correct. In contrast, without using blocks, a drone only predicts the actions of nearby drones within a certain range. For example, D may predict E but not F, while E predicts F. In this case, D knows that E will predict F, and F will also predict some other drones further ahead. Since D does not have complete information about all drones ahead, it cannot make an accurate prediction of E’s actions. 

Without using blocks, even if the prediction accuracy is sufficiently high, another challenge arises from the dependency chain introduced by the prediction process. Each drone's decision depends on the predicted decisions of nearby drones, particularly those ahead of it. Consequently, on a long road with high drone density, the rearmost drone must first predict the decision of the frontmost drone, followed by the second frontmost drone, and so on, before it can determine its own action. As the number of drones increases, this dependency chain substantially increases the computational complexity and results in many unnecessary calculations for drones that are too far away to have any meaningful influence on the current drone's decision.

The use of blocks reduces the computational complexity and breaks the chain of dependence: drones only simulate neighbors within the same block and treat all other drones as conservatively as in the coordination-free approach. In addition, within one block it is easy to establish a priority system based on position and MAC address (to break the tie). Drones near the front boundary of a block (such as E) will make keep-current-lane decisions similar to the previous method. However, drones behind them (like D) in the same block can correctly simulate these decisions and determine that they have higher priority than other competitors, and that the current situation is safe enough to start lane switching. Drones further behind D will also move into the inner lane if the available gap is sufficient and they win the priority. In this way, road usage is distributed more evenly across the lanes.

\subsection{Selection of Block Length}

In the proposed algorithm, a block is introduced to coordinate lane-switching decisions in crowded traffic conditions, where drones cannot determine whether neighboring drones are also attempting to switch to the same lane. Drones within the same block predict other drones' decisions and determine their own actions accordingly. The block length is therefore a critical parameter, as it directly affects the computational complexity, efficiency, and prediction accuracy of the algorithm.

Let $n$ denote the number of neighboring drones detected by drone $A$. In the original STDG algorithm, the computational complexity is $O(n)$, as drone $A$ evaluates each neighboring drone once when calculating the cost function and making the decision. In PB-STDG, drone \(A\) may need to predict the decisions of multiple neighboring drones within the same block before making its own decision. The computational complexity then increases to $O(mn)$, where $m$ is the number of neighboring drones located within the same block. Since $m$ is directly determined by the block length, increasing the block length also increases the computational complexity of the algorithm.

Within each block, drones closer to the front boundary are given higher priority when making decisions. However, these drones may need to consider neighboring drones located in the next-ahead block, whose future actions cannot be predicted. Hence, these neighboring drones are treated as unpredictable potential collision risks. With the additional shared information, including the target lane and the scheduled time of the next algorithm execution, drones can identify and filter out some of these potential collision risks, thereby reducing unnecessary conservatism during the decision-making process. But to ensure safety, the drones at the front of a block may still adopt conservative actions\footnote{However, they also ignore drones that will not execute the algorithm within a short period based on the information shared in the beacon packet. Therefore, even in such cases, the prediction-based algorithm performs more lane switching than STDG.} similar to those in the original STDG algorithm, often slowing down or stopping unnecessarily. The overall efficiency of the algorithm is therefore influenced by the proportion of drones affected by this effect. Increasing the block length decreases the number of blocks and therefore reduces the proportion of drones that may make conservative decisions because of neighboring drones in next-ahead blocks.

In the previous STDG and the proposed algorithm, $\epsilon_1$ defines the distance threshold used to determine whether a neighboring drone is considered a direct or potential collision risk. If another drone is within $\epsilon_1$ of the current drone, it is treated as a direct collision risk if it is in the candidate lane\footnote{The lane that the current drone is considering switching to.}. If it is in a lane adjacent to the candidate lane, it is instead treated as a potential collision risk. A block length satisfying $\epsilon_b < \epsilon_1$ may cause neighboring drones within the potential collision range to be excluded from the prediction process. This is more likely to occur under high-density traffic conditions since drones are closely spaced. Therefore, the prediction-based algorithm may make overly aggressive decisions and lead to an increased collision rate. To ensure that all relevant neighboring drones are considered during prediction, the block length should satisfy
\[
\epsilon_b \geq \epsilon_1.
\]

The final factor influenced by the block length is the prediction accuracy. In the proposed algorithm, drones predict the decisions of neighboring drones based on the information learned from received beacon packets. To achieve accurate predictions, the information available to the predicting drone should be as consistent as possible with the actual information available to the predicted drone. The worst-case communication requirement occurs when the rearmost drone in the current block needs to predict the frontmost drone in the same block. To ensure that the rearmost drone can receive the same safety information as the frontmost drone, the distance can be as large as twice the block length. Since the rearmost drone should at least learn the information for all drones in the next-ahead block, the block length should satisfy:
\[\epsilon_b < \frac{communication\ range}{2}\]

Combining the above considerations, the block length should be carefully selected. It should not be too large, as this increases the computational complexity and reduces prediction accuracy, nor should it be too small, as this may cause the prediction process to ignore nearby drones. Based on this trade-off, the default block length used in this paper is set to 50 m, which is larger than the appropriate value of 15 m for $\epsilon_1$ identified in the previous study and smaller than half of the communication range ($\approx 120$ m). The performance of other block lengths is also evaluated in the simulations to validate this parameter selection.

Based on the assumption that the block length $\epsilon_b$ is set to an appropriate value, we restrict the prediction-based algorithm to consider only drones in the current block and adjacent blocks. This restriction reduces the computational complexity while retaining the information most relevant to short-term collision avoidance and prediction.

% ==========================================================================================

\newcommand{\bstart}[1]{{\ensuremath{\mbox{\textsc{bstart}}(#1)}}}
\newcommand{\bend}[1]{{\ensuremath{\mbox{\textsc{bend}}(#1)}}}

\subsection{Notations}
\label{sec:algorithm:notations}

We inherit the same notation (listed in Table \ref{tab:notation}) as in \cite{qu2026stdg} for consistency, and we additionally define the following symbols used in PB-STDG:
\begin{itemize}
    \item The block length $\epsilon_b$.
    \item The block identifier $b$ represents the b-th block of the road.
    \item The set $H(b)$ contains all existing drones in block $b$ on the road.
    \item For a block $b$, $\bstart{b}$ and $\bend{b}$ denote the arc-length parameters of the start and end points of block $b$, respectively.
\end{itemize}

\begin{table}[htbp]
\centering
\footnotesize
\setlength{\tabcolsep}{4pt}
\renewcommand{\arraystretch}{0.9}
\begin{tabular}{ll}
\toprule
\textbf{Symbol} & \textbf{Description} \\
\midrule
$(i,j)$ & Lane identifier \\
$\neighlanes{i}{j}$ & One-hop neighbor lanes of $(i,j)$ (excluding $(i,j)$) \\
$\candlanes{i}{j}$ & Candidate lanes from $(i,j)$ ($\neighlanes{i}{j} \cup (i,j)$)\\
$\drones{i}{j}$ & Set of drones on lane $(i,j)$ known to the current drone \\
$\dronesunion{\mathcal{L}}$ & Union of $\drones{i}{j}$ for all $(i,j)\in \mathcal{L}$ \\
$\hopdistance{(i,j)}{(i',j')}$ & Hop distance between two parallel lanes $(i,j)$ and $(i',j')$ \\
\dpos{A} & Current arc-length-parameter of drone A \\
$s$ & An arc-length parameter point \\
$C_{max}$ & High cost for marking infeasible candidate lanes \\
$t_{normal}$ & Time interval between two consecutive algorithm executions in normal cases \\
$t_{stop}$ & Time interval between two consecutive algorithm executions when a drone is stopping \\
$t_{s}$ & Time required for a complete lane switch \\
\parconv{\gamma_{i',j'}}{\gamma_{i,j}}{s} & Distance along lane $(i,j)$ obtained by mapping the arc-length parameter $s$ from lane $(i',j')$ \\
\bottomrule
\end{tabular}
\caption{Notation used in the coordination-free STDG algorithm}
\label{tab:notation}
\end{table}

%  ==================================================================================
\subsection{Algorithm Inputs}
\label{sec:algorithm:inputs}

The algorithm inputs are the neighbor table constructed from received beacon packets, the drone road system map, and the current drone's own information. For each drone $A$ in the neighbor table on the same road as the current drone (including the current drone itself), we collect:
\begin{itemize}
    \item The drone ID/MAC Address $A$.
    \item The current speed $v^A$ and the preferred speed $v_{p}^A$.
    \item The current arc-length parameter $s^A$.
    \item The current lane identifier $(i,j)^A$.
    \item The target lane identifier $(i_T, j_T)^A$, associated arc-length parameter point $s_{(i_T, j_T)^A}$, and the distance $d_{(i_T, j_T)^A}^{target}$ from the current position to that point. These values are set to None if drone $A$ does not have a target lane.
    \item The lane $(i_S,j_S)^A$ drone $A$ is currently switching to; it is set to None if drone $A$ is not currently switching lanes.
    \item The time $t_n^A$, which represents the remaining time until drone \(A\) executes its next invocation of the algorithm. It is calculated by subtracting the current time from the next algorithm time recorded in the beacon.
\end{itemize}

Here, we pre-process these inputs into the following forms for use in subsequent algorithm steps. This pre-processing stage is executed at the start of every algorithm invocation.

\begin{itemize}
\item The current block identifier $b$ of the current drone $A$. This is defined as
\begin{equation}
    b^A =
    \left\lfloor
    \frac{\parconv{\gamma_{(i,j)^A}}{\gamma_{i_0,j_0}}{s^A}}{\epsilon_b}
    \right\rfloor
\end{equation}
where $\epsilon_b$ is a configurable parameter that determines the block length used to split a road into several equal-sized blocks, and $(i_0,j_0)$ is the central lane identifier.

\item The set of competitor drones $O(A)$ of the current drone $A$ is defined as
\begin{equation}
\label{eq:competitor}
    O(A) =
\left\{
A^\prime \;\middle|\;A^\prime \in H(b^A),\;t^{A^\prime}_n < t_c,\;(i_S,j_S)^{A^\prime} = \varnothing
\right\}
\end{equation}
where $t_c$ is an algorithm parameter used to determine the drones that may potentially collide with the current drone. If $t_n^{A'} < t_c$, this means that another drone $A'$ will start its algorithm within a very short time of the current drone $A$, and may suffer a concurrent collision. These competitors should be in the same block as the current drone and should not be switching lanes.

\item For each lane $(i,j)$ in the current road, the set of direct hazard drones $D(i,j)^A$ is
\begin{equation}
\begin{aligned}
D(i,j)^A = \Big\{
A^\prime \;\Big|\;&
A^\prime \in \bigcup_{k=-1}^{1} H(b^A+k), \\
&( (i,j)^{A^\prime} = (i,j) )
\;\lor\;
((i_S,j_S)^{A^\prime} = (i,j)), \\
& s^{A^\prime} \in
\Big[
\bstart{b^A}-\epsilon_1,\;
\bend{b^A}+\epsilon_1
\Big]
\Big\}
\end{aligned}
\end{equation}
A direct hazard drone is a drone with which other drones may collide if they directly switch to its lane. A drone $A^\prime$ is considered a direct hazard of lane $(i,j)$ if it is either currently on that lane or is switching to that lane. And it should be within a “buffered” range $\epsilon_1$ around block $b^A$.

\item For each lane $(i,j)$ in the current road, the set of potential hazard drones $P(i,j)^A$ is
\begin{equation}
\begin{aligned}
P(i,j)^A = \Big\{ A' \;\Big|\;&
A' \in \bigcup_{k=0}^{1} H(b^A+k), \\
& \hopdistance{(i,j)^{A^\prime}}{(i,j)}=1,\; t^{A'}_n \le 0 \;\lor \\
& \Big( t^{A'}_n < t_c \;\wedge\; 
s^{A^\prime} \in \Big[
\bend{b^A},\;
\bend{b^A}+\epsilon_1
\Big]
\Big)
\Big\}
\end{aligned}
\end{equation}
A potential hazard drone is one whose behavior is highly uncertain and may result in a collision with other drones switching to the same lane simultaneously. A drone $A^\prime$ can be seen as a potential hazard of lane $(i,j)$ if its current lane $(i,j)^{A^\prime}$ is a neighbor of $(i,j)$, and it either has a negative value\footnote{This is caused by packet loss or delays in the beacon packets.} of $t^{A'}_n$ or is in the next block of $b$ with $t^{A'}_n < t_c$. The selection areas corresponding to the sets $O(A)$, $D(i,j)^A)$, and $P(i,j)^A$ are shown in Figure \ref{fig:selection area}. 

\begin{figure}[htbp]
  \centering
  \includegraphics[scale=0.5]{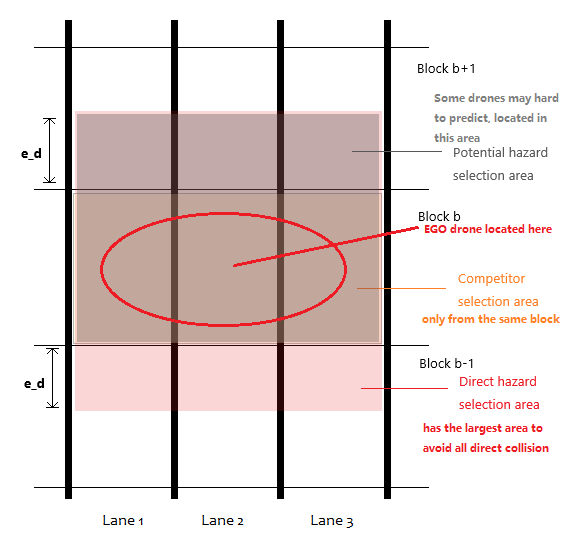}
  \caption{The selection areas of drones in $O(A)$, $D(i,j)^A)$, and $P(i,j)^A$. Drones are assumed to receive complete information from at least the current block and blocks immediately ahead of and behind that.}
  \label{fig:selection area}
\end{figure}

\item The set $C((i,j)^{A^\prime})^A$ of candidate
  lanes for each drone $A^\prime$ in set $O(A) \cup \{A\}$. For each lane 
  $(i',j')^{A^\prime}\in C((i,j)^{A^\prime})^A$ the following information is further collected:
  \begin{itemize}
  \item The distance-along-the-lane $d_{(i',j')^{A^\prime}}^{ahead}$ to the closest drone ahead on lane $(i',j')^{A^\prime}$.
  \item The speed $v_{(i',j')^{A^\prime}}$ of the closest drone ahead on lane
    $(i',j')^{A^\prime}$. It is assigned a value of infinity when there is no drone ahead.
  \item The number of direct hazard drones $\Delta^{\mbox{direct}}_{(i',j')^{A^\prime}}$ for drone $A'$ is defined as:
    \begin{equation}
       \Delta^{\mbox{direct}}_{(i',j')^{A^\prime}} = \magnitude{ \setof{
       A^{\prime\prime} \in D((i',j')^{A^\prime})^A}
       {|\parconv{\gamma_{i'',j''}}{\gamma_{i',j'}}{s^{A'}} - \dpos{A''}| < \epsilon_1}  
       }
    \end{equation}
  And the number of potential hazard drones $\Delta^{\mbox{potential}}_{(i',j')^{A^\prime}}$ drone $A'$ is defined as:
    \begin{equation}
       \Delta^{\mbox{potential}}_{(i',j')^{A^\prime}} = \magnitude{ \setof{
       A^{\prime\prime} \in P((i',j')^{A^\prime})^A}
       {|\parconv{\gamma_{i'',j''}}{\gamma_{i',j'}}{s^{A'}} - \dpos{A''}| < \epsilon_1}  
       }
    \end{equation}
where $\epsilon_1 > 0$ is a threshold parameter used to control the selection range, and $(i'',j'')$ denotes the lane occupied by drone $A''$.

These two numbers are used to represent the risk of the candidate lane.
  
  \item The achievable speed $v^*_{(i',j')^{A^\prime}}$ of drone $A'$ candidate
    lane $(i',j')^{A^\prime}$, it is
    {\small
    \begin{equation}
      \label{eq:candidate-achievable-speed}
      v^*_{(i',j')^{A^\prime}} = \left\{\begin{array}{r@{\quad:\quad}l}
              v_p^{A'} & (d_{(i',j')^{A^\prime}}^{ahead} > \epsilon_2) \wedge \big((i',j')^{A^\prime} = (i_T, j_T)^{A'}\big) \\
        \min\set{v_p^{A'}, d_{(i_T, j_T)^{A'}}^{target}/t_{s}} & (d_{(i',j')^{A^\prime}}^{ahead} > \epsilon_2) \wedge \big((i',j')^{A^\prime} \neq (i_T, j_T)^{A'}\big) \\
        \min\set{v_p^{A'}, v_{(i',j')^{A^\prime}}, d_{(i_T, j_T)^{A'}}^{target}/t_{s}} & (d_{(i',j')^{A^\prime}}^{ahead} \le \epsilon_2) \wedge \big((i',j')^{A^\prime} \neq (i_T, j_T)^{A'}\big) \\
                \min\set{v_p^{A'}, v_{(i',j')^{A^\prime}}} & (d_{(i',j')^{A^\prime}}^{ahead} \le \epsilon_2) \wedge \big((i',j')^{A^\prime} = (i_T, j_T)^{A'}\big)\\
      \end{array}
      \right.
    \end{equation} 
    }
    The achievable speed is the minimum of the preferred speed, the speed of the drone ahead (if it exists and its distance-along-the-lane from drone $A'$ is less than a threshold $\epsilon_2$), and the distance to the target point divided by the switching time $t_s$.
  \end{itemize}
\item For drone $A'$ belongs to $O(A)\cup \{A\}$, the set of restricting lanes $B((i,j)^{A^\prime})^A \subset C((i,j)^{A^\prime})^A \setminus \{(i_T, j_T)^{A'}\}$. These lanes include the attachment points to non-target connection or off-ramps of drone $A'$ and may likely be occupied by other drones targeting one of these ramps.
\end{itemize}

% ==================================================================================
\subsection{Cost Model}
\label{sec:algorithm:costModel}

In PB-STDG, we need to calculate the total cost for the current drone and the competitor drones belonging to $O(A)$. Thus the equation of total cost for drone $A'$ in a candidate lane $(i',j')^{A'}$ is:
\begin{equation}
    c((i',j'), v^*_{(i',j')^{A'}})^{A'} = c_s(v^*_{(i',j')^{A'}})^{A'}
  + \kappa_1 \cdot c_p(i',j')^{A'}
  + \kappa_2 \cdot c_c(i',j')^{A'}
\end{equation}
where $\kappa_1, \kappa_2 \ge 0$ are non-negative weight parameters.

In the total cost function:
\begin{itemize}
    \item $\fnct{c_s}$ is the speed cost function which assigns a cost related to the speed loss (i.e., the difference between the preferred and achievable speed) on that lane. It is defined as
    \begin{equation}
         c_s(v)^{A'} = \left(v_p^{A'} - v\right)^2
    \end{equation}
    This model imposes a stronger penalty for larger speed loss.

    \item $\fnct{c_p}$ is the position cost function that quantifies how urgently the drone should move toward its target lane or away from the target lanes of other drones. It is defined as
    {\small
    \begin{equation}
      c_p(i',j')^{A'} = \left\{\begin{array}{r@{\quad:\quad}l}
        f(d_{(i_T, j_T)^{A'}}^{target}) \cdot \hopdistance{(i',j')}{(i_T,j_T)^{A'}}
              & ((i_T,j_T)^{A'}) \ne \mbox{none}) \wedge((i',j') \notin B((i',j')^{A^\prime})^A) \\
        C_{max} & ((i',j') \ne ((i_T,j_T)^{A'}) \wedge ((i',j') \in B((i',j')^{A^\prime})^A)\\
        C_f-\hopdistance{(i',j')}{(i_f,j_f)} & ((i_T,j_T)^{A'}) = \mbox{none}) \wedge((i',j') \notin B((i',j')^{A^\prime})^A)      \\
        0      & \mbox{otherwise}
      \end{array}
      \right.
    \end{equation}}
    Here, $f(d_{(i_T, j_T)^{A'}}^{target})$ is a non-negative decreasing function that provides a multiplier representing the urgency level for the position cost if the drone $A'$ has a target lane. The influence of the position cost is small when the drone is far from its target. In this case, the term $\hopdistance{(i',j')}{(i_T,j_T)^{A'}}$ represents a cost penalty that decreases as the hop distance from the candidate lane to the target lane becomes smaller. When drone $A'$ has no target lane, candidate lanes leading to other ramps are assigned the maximum cost penalty. For other candidate lanes $(i',j')$, the cost penalty is defined as a constant $C_f$ minus the hop distance to the farthest lane $(i_f,j_f)$ that includes an attachment point to another ramp. Otherwise, the cost is set to zero.

    \item $\fnct{c_c}$ is the collision cost function, which quantifies the risk associated with switching to the candidate lane. It is defined as
    \begin{equation}
      c_c(i',j')^{A'} = \left\{\begin{array}{r@{\quad:\quad}l}
        C_{max} & ((i',j')\ne(i,j)^{A'}) \wedge (\Delta^{\mbox{direct}}_{(i',j')^{A^\prime}}>0) \\
        0 & (i',j') = (i,j)^{A'} \\
        \Delta^{\mbox{potential}}_{(i',j')^{A^\prime}} & \mbox{otherwise}
      \end{array}
      \right.  
      \label{eq:collision_cost}
    \end{equation}
    Any positive value of $\Delta^{\mbox{direct}}_{(i',j')^{A^\prime}}$ indicates a high risk of direct collision with other drones in lane $(i',j')$, thus the cost is set to the maximum value. If the candidate lane is the current lane, no switching occurs and there is no risk of collision in this case. Otherwise, the cost equals the value of $\Delta^{\mbox{potential}}_{(i',j')^{A^\prime}}$, reflecting the number of drones with uncertain behavior that could concurrently switch to the candidate lane and induce a collision.
    
\end{itemize}

% ==================================================================================
\subsection{Predict Competitor Decisions and Update Metrics}
\label{sec:algorithm:decision}

After obtaining the algorithm inputs, a competitor set $O(A)$ is constructed. The next step is to predict the decisions of these competitors using the information contained in their beacon packets. To ensure that all drones predict their competitors in the same order, a priority comparison function $p_d(A,A')$ is introduced to determine the relative priority between two drones. In this paper, we have
\begin{equation}
\label{eq:dronePriority}
    p_d(A,A') = (s^{A} > \parconv{\gamma_{(i,j)^A}}{\gamma_{(i,j)^{A'}}}{s^{A'}}) \;\lor\; (s^A = \parconv{\gamma_{(i,j)^A}}{\gamma_{(i,j)^{A'}}}{s^{A'}} \;\wedge\; (A>A') )
\end{equation}

We only consider competitors whose priority is higher than that of the current drone $A$, since their decisions may affect drone $A$. Competitors with lower priority than $A$ are not predicted by $A$; instead, they are required to take the decision of $A$ into account when making their own decisions.

In this step, the algorithm iteratively selects the yet-unpredicted competitor $A'$ with the highest priority from the set. If its priority is lower than that of the current drone, or if the set is empty, the iteration terminates and the final decision for the current drone is made. Otherwise, the following procedure is executed:

\begin{enumerate}
  \item  For each candidate lane 
  $(i',j')^{A^\prime}\in C((i,j)^{A^\prime})^A$ of drone $A'$, calculate the weighted total cost $c((i',j'), v^*_{(i',j')^{A'}})^{A'}$.
  
  \item Select the predicted output pair $((i_o, j_o)^{A'}, v_{(i_o,j_o)^{A'}})$ based on the following equations
    \begin{equation}
    \label{eq:getmin}
    c_{\min}^{A'} \triangleq 
    \min_{((i',j'),\, v^*_{(i',j')^{A'}})\in \tau}
    c\big((i',j'), v^*_{(i',j')^{A'}}\big)^{A'}
    \end{equation}
    
    \begin{equation}
    \label{eq:getoutput}
    \big((i_o, j_o)^{A'},\, v_{o}^{A'}\big) =
    \begin{cases}
    \displaystyle
    \argmin_{((i',j'),\, v^*_{(i',j')^{A'}})\in \tau}
    c\big((i',j'), v^*_{(i',j')^{A'}}\big)^{A'}, 
    & c_{\min}^{A'} < C_{\max}, \\[6pt]
    \big((i,j)^{A'},\, 0\big),
    & c_{\min}^{A'} \ge C_{\max}.
    \end{cases}
    \end{equation}
  where $\tau$ denotes the set of candidate lane--speed pairs. 
  
  In the original STDG algorithm, when multiple candidate lanes have the same minimum cost below $C_{\max}$, a lane is randomly selected from these candidates as the output lane. In the prediction-based algorithm, however, this selection process needs to be deterministic and known to all drones so that they obtain the same result. Therefore, a lane priority function $p_l!\big((i,j),(i',j')\big)$ is introduced to break ties. In this paper, we adopt a simple form:
  \begin{equation}
      p_l\!\big((i,j),(i',j')\big) = (i*k+j) > (i'*k+j') 
  \end{equation}
  The constant $k$ is chosen as a sufficiently large integer to ensure that different pairs $(i,j)$ output distinct values of $(i*k + j)$.
  If the current lane $(i,j)^{A'}$ also outputs the minimal cost, we directly select $(i_o, j_o)^{A'} = (i,j)^{A'}$ without applying the lane priority comparison. Otherwise, the lane with the highest priority is selected.

  \item If the output lane coincides with the current lane but differs from the target lane, determine whether the competitor needs to stop in order to avoid missing the target point by the following criterion:
    \begin{equation}
    \label{eq:check_miss_switchPoint}
      v_{o}^{A'} =
      \left\{\begin{array}{r@{\quad:\quad}l}
      0 & (d_{(i_T, j_T)^{A'}}^{target} < \epsilon_2) \wedge (d_{(i_o, j_o)^{A'}}^{ahead} < \epsilon_2) \\
      d_{(i_T, j_T)^{A'}}^{target}/t_{normal}&(d_{(i_T, j_T)^{A'}}^{target} < \epsilon_2) \wedge (d_{(i_o, j_o)^{A'}}^{ahead} \geq \epsilon_2) \\ 
      v_{o}^{A'} & \mbox{otherwise}
      \end{array}
      \right.    
    \end{equation}

  \item Based on the predicted decision, update:
  \begin{enumerate}
      \item $v^{A'}$ to $v_{o}^{A'}$
      \item $(i,j)^{A'}$ to $(i_o, j_o)^{A'}$
      \item add $A'$ to $D(i_o, j_o)^A$ if it does not exist
  \end{enumerate}

  \item Remove $A'$ from $O(A)$

\end{enumerate}

% ==================================================================================
\subsection{Make Final Decision and Schedule the Next Run}
\label{sec:algorithm:s4}

After simulating the decisions that other competitors may make and assuming that they have acted according to the predicted decisions, the algorithm proceeds to make the final decision for the current drone $A$ by applying the same steps from 1 to 3 described in Section \ref{sec:algorithm:decision}. The current drone $A$ will then switch to the output lane with the output speed, and immediately broadcast a beacon to inform nearby drones of its updated state.

The period until the next algorithm invocation $t_{next}$ depends on the final decision. If the output lane differs from the current lane, $t_{next}$ is set to $t_{s}$ to ensure that the drone can adjust immediately upon completing the lane switch. Otherwise, if the output speed is nonzero, $t_{next} = t_{normal}$; if the output speed is zero, $t_{next} = t_{stop}$ where $t_{stop} < t_{normal}$. This setting allows stopped drones to check their surrounding environment more frequently, ensuring a faster exit from the stopped state.

The overall pseudocode is listed below.
\RestyleAlgo{ruled}
\SetKwComment{Comment}{/* }{ */}
\begin{algorithm}[hbt!]
\scriptsize
\SetAlgoNlRelativeSize{0}
\LinesNumbered
\caption{PB-STDG: Prediction-Based Short-Term Decentralized Greedy}\label{alg:stdg}
\KwData{collect algorithm inputs (see Section \ref{sec:algorithm:inputs})}
\KwResult{$(i_o,j_o), v_o$}
$v_o \gets 0$; $(i_o,j_o) \gets (i,j)$\;
$O(A) \gets $ updated and sorted according to Eq.\eqref{eq:competitor} and \eqref{eq:dronePriority} \;
\For{$A' \in O(A)$}{
    \If{$O(A)$ is empty or $p_d(A,A')$ is True} {break\;}
    $v_o^{A'} \gets 0$\;
    $(i_o,j_o)^{A'} \gets (i,j)^{A'}$ \;
    $c_o^{A'} \gets C_{max}$\;
    $C((i,j)^{A'})^A \gets \neighlanes{i}{j}^{A'} \cup \set{(i,j)^{A'}}$\;
    \For{$(i^\prime,j^\prime)^{A'} \in C((i,j)^{A'})^A$}{
    $c_{total} \gets c_s(v^*_{(i',j')^{A'}})^{A'}
  + \kappa_1 \cdot c_p(i',j')^{A'}
  + \kappa_2 \cdot c_c(i',j')^{A'}$\;
    \If{$c_{total}^{A'} < c_o^{A'}$ \textbf{or} ($c_{total}^{A'} == c_o^{A'}$ \textbf{and} ( $p_l\!\big((i',j')^{A'},(i_o,j_o)^{A'}\big)$ \textbf{or} $(i_o,j_o)^{A'}$ == $(i,j)^{A'}$ ) )  }{
      $c_o^{A'} \gets c_{total} $\;
      $v_o^{A'} \gets v^*_{(i',j')^{A'}}$\;
      $(i_o,j_o)^{A'} \gets (i',j')^{A'}$\;
    }
    }
    Update algorithm metrics based on step 4 described in Section     \ref{sec:algorithm:decision}\;
    Remove $A'$ from $O(A)$
    
}
Make own decision: $v_o,(i_o,j_o)\gets$ updated using the same procedure as in Lines 7–18 by replacing $A'$ with $A$\;
Switch to lane $(i_o, j_o)$ with speed $v_o$\;
Schedule the next execution of algorithm based on the method introduced in Section \ref{sec:algorithm:s4}
\end{algorithm}

% ==========================================================================================
%                        Simulation Results
% ==========================================================================================
\section{Simulation Results}
\label{sec:results}

In this section, we present several simulation results to compare the performance metrics, mainly the collision rate and average speed, between the STDG and prediction-based algorithms. We explore the impact of environmental parameters and algorithm configurations, and conduct a sensitivity study. Finally, we discuss this algorithm and outline directions for future research.

% ==========================================================================================
\subsection{Simulation Environment and Assumptions}
\label{sec:results:environment-and-assumptions}

We run the simulations on OMNeT++ 5.6.2 with the INET framework version 4.2.5. Table \ref{tab:simulation-parameters} lists the fixed simulation and algorithm parameters.

\begin{table}[ht!]
  \centering
  \renewcommand{\arraystretch}{0.8}
\footnotesize
  \begin{tabular}{|l|l|}
    \hline
    \textbf{Parameter}                  & \textbf{Value}                     \\ \hline
    Path Loss Exponent                  & 2                                  \\ \hline
    Minimal Safety Distance             & 0.5 m                              \\ \hline
    Lane Radius                         & 10 m                            \\ \hline
    Preferred Speed Interval                      & 10--15 m/s                         \\ \hline
    Preferred Speed Distribution                  & Uniform                            \\ \hline
    Replication Number per Simulation   & 50                                 \\  \hline
    Simulation time                     & 500 s                             \\
    \hline
    Algorithm Period (Normal)        $t_{normal}$    & 2 s     \\ \hline
    Algorithm Period (Stop)       $t_{stop}$     & 0.5 s     \\ \hline      
    Time for lane switching        $t_s$    & 1 s                                \\ \hline
        Time Threshold       $t_c$    & 0.5 s                                \\ \hline
    Maximum Cost Definition         $C_{max}$    & 30000                  \\ \hline
    Distance Threshold $\epsilon_2$   &  30 m  \\
    \hline
  \end{tabular}
\caption{Fixed Simulation Parameters}
  \label{tab:simulation-parameters}
\end{table}

In the prediction-based algorithm, we introduce an additional parameter, block length $\epsilon_b$. In Section \ref{sec:algorithm}, we introduced guidelines for selecting a suitable value of $\epsilon_b$. In our simulations, this parameter is fixed at 50 meters for most scenarios. 

During the simulation process, we make the following assumptions:
\begin{itemize}
    \item Drones are assumed to have knowledge of the tested DRS. Once a drone is injected into the system, it is assigned a random destination. The end-to-end path from its start point to the destination is also known by the drone, and it will follow that path during its whole lifespan.
    \item Drones in the road system are treated as points. A collision is considered to occur when the distance between two drones is smaller than the minimum safety distance. The two drones involved in the collision are immediately removed from the simulation and no longer affect other drones in the system. However, their safety information is still retained by other drones until no packet has been received from them for ten seconds.
    \item For each lane with an entry point at the beginning of the road, drones are continuously injected into the lane at randomly determined time intervals. The inter-arrival time is drawn from an exponential distribution with a mean determined by the $Generation\ Rate$. The generation rate represents the ideal number of drones generated per second for each lane. However, when the road system becomes fully occupied by drones, additional drones cannot be injected. In such a case, the actual number of drones injected into the system per second per lane is referred to as the $Injection\ Rate$, which can be much lower than the $Generation\ Rate$ under high-density conditions.
    \item Once a drone reaches its destination, it is immediately removed from the simulation, but its information remains in other drones' neighbor tables until the corresponding entries expire. The neighbor table stores all safety information received from other drones, and its entries are maintained as soft state.
    \item Drones accurately know and send their current position to neighbors. GPS noise is ignored during the simulation.
    \item In the original STDG algorithm, a threshold of $2\epsilon_1$ is used to identify potential risks in neighboring lanes, whereas the prediction-based algorithm uses a threshold of $\epsilon_1$. To provide a consistent basis for comparison, a threshold of $\epsilon_1$ is used for both algorithms in all subsequent simulations. This adjustment increases the average speed of STDG, while its collision rate remains close to that obtained with the original threshold.
\end{itemize}

The key performance metrics are \textbf{Collision Rate (cr)}, which is the fraction of collided drones among all injected drones; \textbf{Average Speed (as)}, which indicates the average speed of all injected drones; and \textbf{Throughput}, which is the number of drones arriving at their destinations per second, also referred to as the arrival count. With low collision rates, the throughput should be practically the same as the injection rate. The preliminary study shows that the system reaches a steady state after an initial transient period. The detailed period depends on the road system, and the largest value among all tested road systems is 200 seconds. Thus we only count the throughput after that time in all simulations. Additional metrics will also be investigated for specific purposes in the following sections. For all following simulations, we run 50 replications and the presented results are the average values for these replications to ensure they are statistically meaningful.

%========================================================================
\subsection{Comparing the Two Algorithms on Simple DRS}
\label{sec:results:initial-study}

\begin{figure}[htbp]
  \centering
  \includegraphics[scale=0.3]{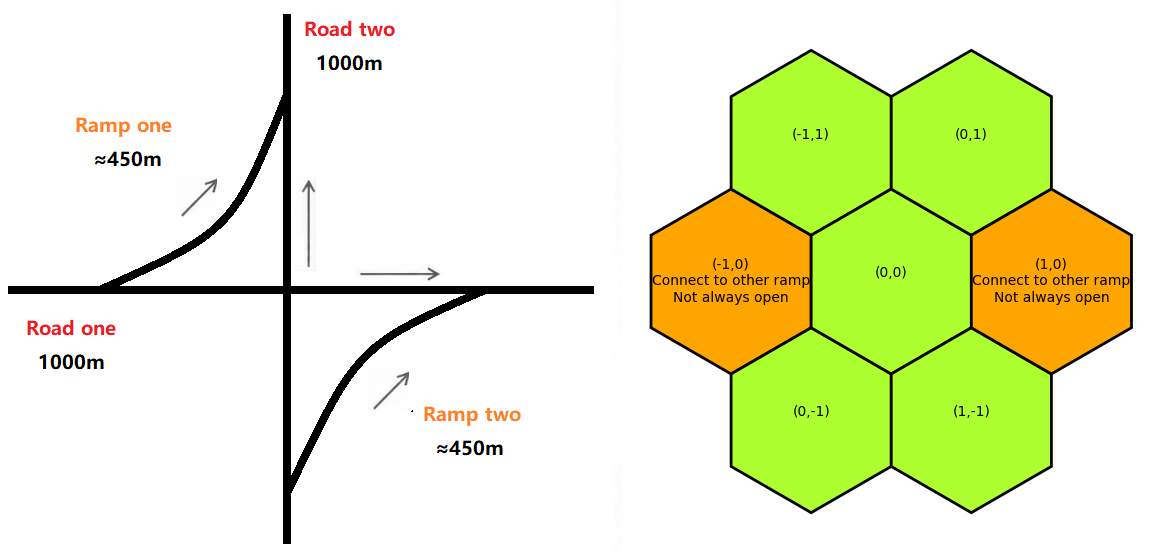}
  \caption{A simple DRS used to compare STDG and PB-STDG. The two roads are located at different altitude levels. The right-hand side shows the lane arrangement of both roads.}
  \label{fig:2r2r}
\end{figure}

First, a preliminary study was conducted to explore the impact of MAC/PHY parameters, including transmit power and beaconing rate. The tested values were 2 mW and 5 mW for transmit power, and 5 Hz, 10 Hz, 20 Hz, and 50 Hz for beaconing rate. Both the STDG and prediction-based algorithms were evaluated on a simple road system consisting of two roads connected by two ramps (shown in Figure \ref{fig:2r2r}). This simple DRS allows us to evaluate the ability of the algorithm to help drones reach their destinations, overtake slower drones, switch to another road, and avoid collisions.

Table \ref{tab:algorithm-parameters-inital-study} lists the value of key algorithm parameters used in this section. We choose a conservative position-to-collision cost weight ratio, $\kappa_1/\kappa_2$. A small value makes the total cost more likely to reach $C_{max}$, causing drones to make more conservative decisions. The distance threshold is also selected to ensure close-to-zero collisions will happen if no packet loss occurs. Therefore the effects of the MAC/PHY parameters on the algorithms could be clearly observed. 
\begin{table}[ht!]
  \centering
  \renewcommand{\arraystretch}{0.8}
  \begin{tabular}{|l|l|l|}
    \hline
    \textbf{Parameter}              & \textbf{Notation}    & \textbf{Value}                     \\ \hline
    Position Cost Coefficient         & $\kappa_1$    & 1                  \\ \hline
    Collision Cost Coefficient         & $\kappa_2$    &  10000             \\ \hline
    Distance Threshold for Searching Neighboring Drones &   $\epsilon_1$    &  15 m  \\ \hline
  \end{tabular}
  \caption{Key Algorithm Parameters in Initial Study}
  \label{tab:algorithm-parameters-inital-study}
\end{table}

The results show that there are no significant differences between different transmit power–beaconing rate combinations, except for all 50 Hz cases, which result in a 0.2\% collision rate across all transmit powers and algorithms, compared to the close-to-zero collision rates observed in the other cases. This result is consistent with the previous study of the STDG algorithm. Therefore, in all subsequent simulations, 2 mW and 10 Hz are selected as the fixed values for transmit power and beaconing rate.

During the exploration of the STDG algorithm, one finding was that an extremely small $\kappa_1/\kappa_2$ ratio (e.g., 1/10000) results in the lowest collision rate, but also the lowest average speed compared to other ratios such as 1/1, 1/100, and 100/1. Since the objective of the proposed prediction-based algorithm is to improve the average speed without degrading the collision rate, the following simulations in this section focus on comparing the average speed and throughput under the $\kappa_1/\kappa_2 = 1/10000$ setting.

\begin{figure}[htbp]
  \centering
  \includegraphics[scale=0.5]{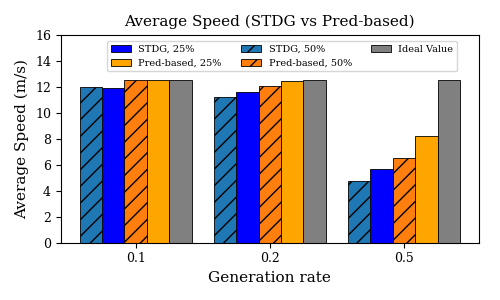}
  \caption{Average speed for STDG and Prediction-Based algorithms in two-road two-ramp scenario, where the beaconing rate is 10 Hz and the transmit power is 2 mW.}
  \label{fig:2r2r_as}
\end{figure}

From the results, the average collision rates for all algorithms and all varied parameters are always close-to-zero (average value $<0.007\%$, with the maximum value 0.17\% over all replications). Figure \ref{fig:2r2r_as} shows the average speed achieved by the STDG and prediction-based algorithms. The gray bars are fixed at 12.5 m/s, which represents the ideal average speed calculated from the default uniform speed distribution ranging from 10 m/s to 15 m/s. The percentage values shown in the legend (25\% and 50\%) represent the proportion of drones whose destinations are different from their initial roads. Since the evaluated scenario consists of two roads connected by two ramps, a value of 50\% indicates that half of the drones must cross a connecting ramp to move from their spawn road to the other road, and the remaining half continue cruising on their original roads throughout the simulation.

The most obvious finding is that the prediction-based algorithm achieves a higher average speed than the STDG algorithm in all cases. As the generation rate increases, the differences between the two algorithms become more noticeable. When the generation rate ($gr$) is 0.1 or 0.2, the average speeds achieved by the prediction-based algorithm can even reach the ideal value. When $gr = 0.5$, the average speeds of both algorithms decrease substantially but the performance gap between the two algorithms becomes more pronounced, with values of 4.71 m/s versus 6.51 m/s in the 50\% case, and 5.70 m/s versus 8.19 m/s in the 25\% case for the STDG and prediction-based algorithms, respectively. Another observation is that the average speed decreases as the proportion of drones whose entry and exit roads are different increases. This trend becomes more evident at higher generation rates.

\begin{figure}[htbp]
  \centering
  \includegraphics[scale=0.5]{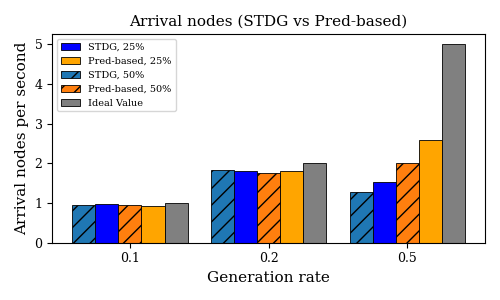}
  \caption{Arrival count for STDG and Prediction-Based algorithms in two-road two-ramp scenario, where the beaconing rate is 10 Hz and the transmit power is 2 mW.}
  \label{fig:2r2r_ac}
\end{figure}

Figure \ref{fig:2r2r_ac} shows the throughput for the same simulations. Since each of the two tested roads consists of five lanes, the ideal arrival count is equal to the total number of lanes (i.e., 10) multiplied by the generation rate. Then, the ideal arrival counts are 1, 2, and 5 for generation rates ($gr$) of 0.1, 0.2, and 0.5, respectively. When some roads become congested, the actual arrival count may be much lower than the ideal value because no additional drones can be injected into these roads for a period of time. This is the main reason for the difference between the actual and ideal arrival counts. Since the collision rates are close-to-zero in all cases, the impact of collisions on the arrival count can be largely ignored.

%============================================================

For generation rates ($gr$) of 0.1 and 0.2, the arrival counts per second are similar for both algorithms and are only slightly lower than the ideal values. However, in the $gr = 0.5$ cases, the prediction-based algorithm demonstrates substantially better performance than the STDG algorithm. Even in the 50\% case, where half of the drones need to switch roads, the prediction-based algorithm achieves a higher arrival count than the 25\% case of the STDG algorithm.

\subsection{Evaluation of Lane Usage Distribution}
\label{sec:results:position}

During the study of the STDG algorithm, a hypothesis was proposed regarding the existence of an inner lane vanishing phenomenon. Based on observations from the simulation results, this phenomenon refers to the tendency of inner lanes to become frequently less occupied when the generation rate is high. The prediction-based algorithm considers the inner lane vanishing problem and aims to improve the utilization of inner lanes, with the potential to improve average speed and throughput.

In this section, lane usage distributions are collected and analyzed to evaluate the effects of different algorithms. The usage of lane $(i,j)$ is defined as
\[
usage_{i,j} = \frac{\sum_{k=0}^{3000}f_{i,j}(200+0.1k)}{\sum_{k=0}^{3000}n(200+0.1k)}
\]
where $f_{i,j}(t)$ denotes the number of drones in lane $(i,j)$ at time $t$, and $n(t)$ denotes the total number of drones across all lanes in the same road at time $t$. The simulation reaches a steady state after 200 s and ends at 500 s. Therefore, drone positions are sampled every 0.1 s over the interval from 200 s to 500 s, yielding 3001 samples. The lane usage is then computed as the ratio of the accumulated number of drones in lane $(i,j)$ to the accumulated number of drones across all lanes during the sampling period.  

We first reproduce the previous simulations in which the inner lane vanishing phenomenon was observed under the $\kappa_1/\kappa_2 = 1/10000$ setting. The results show that the inner-lane (i.e. lane (0,0)) usage of STDG is 12.5\% and 14.5\% for generation rates ($gr$) of 0.1 and 0.2, respectively. We then evaluate the prediction-based algorithm using the same parameter settings. The corresponding inner-lane usage increases to 16.8\% and 18.3\%, respectively. The average speeds of STDG and the prediction-based algorithm are 11.9 m/s and 12.4 m/s for $gr=0.1$, and 7.1 m/s and 12.3 m/s for $gr=0.2$, respectively. In this scenario, the prediction-based algorithm improves inner-lane utilization. Many drones that would otherwise be blocked by slower drones ahead are able to switch to the central lane, where they can travel at their preferred speed or experience less speed reduction. Therefore, the average speed is increased.

Next, two additional road systems are introduced to further investigate the effects of the two algorithms on lane usage. One system consists of a single 1000 m long road without any ramps, allowing the effects of the algorithms to be evaluated without bias caused by different destinations. Its lanes form a three-tier hexagonal network, where the first tier contains the central lane with lane identifier (0,0), the second tier consists of six lanes directly connected to the central lane, and the third tier contains twelve lanes connected to the second-tier lanes. Another system is derived from the first system by introducing an off-ramp connected to lane (2,0) at the end of the road to evaluate the algorithms under biased traffic conditions. We refer to them as "1r0r" (1 road and zero ramps) and "1r1r" (1 road with 1 ramp) in the following discussion.

\begin{figure}[htbp]
  \centering
  \includegraphics[scale=0.4]{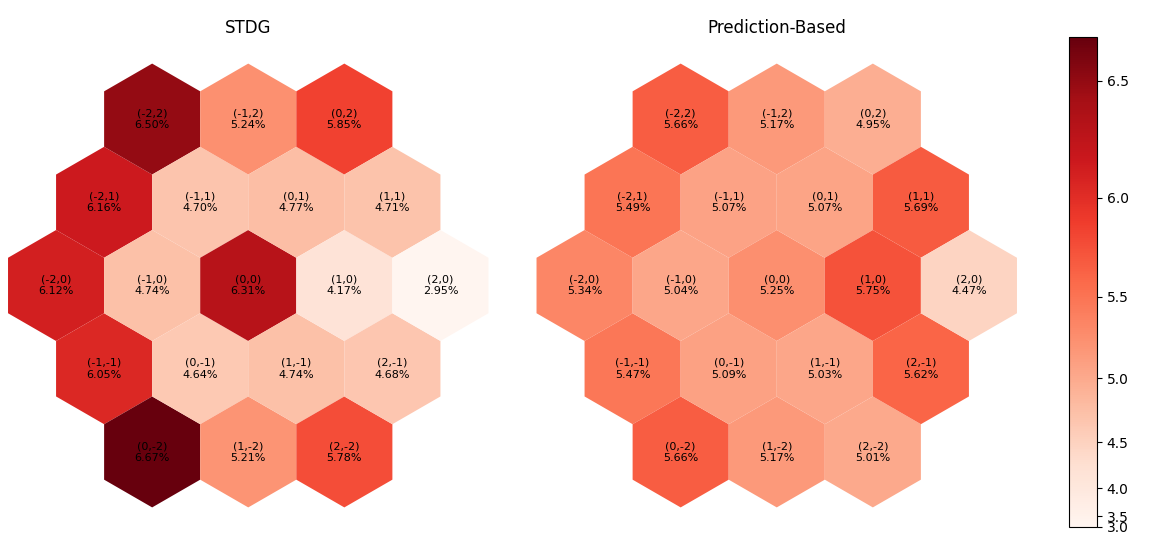}
  \caption{Lane Usage Distributions for STDG and Prediction-Based Algorithms in One Road One Ramp Scenario. In each hexagon, $(i,j)$ denotes the lane identifier, and the percentage indicates the corresponding lane usage.}
  \label{fig:ld_1r1r}
\end{figure}

We set the generation rate to 0.5 for both road systems to simulate a high-density scenario. For the 1r0r case, the simulation results show that no collisions occurred for either algorithm, and the lane usage distributions for them remain uniform, with the percentage for each lane being approximately 5\%. However, in the 1r1r scenario, when half of the drones need to switch to the ramp, a lane vanishing phenomenon similar to that observed previously appears in lane (2,0) when using the STDG algorithm. The prediction-based algorithm also exhibits this phenomenon, although it is much less pronounced. Figure \ref{fig:ld_1r1r} shows the lane usage distributions for both algorithms.

For the STDG algorithm, lane (2,0) exhibits the lane vanishing phenomenon. Its lane usage of 2.95\% is substantially lower than that of all other lanes. In contrast, the five left-hand-side lanes ((-2,2), (-2,1), (-2,0), (-1,-1), and (0,-2)), together with the central lane (0,0), carry a substantially higher proportion of drones. Meanwhile, the lanes adjacent to (0,0) and (2,0) also exhibit relatively low lane usage, although the imbalance is less pronounced than that of lane (2,0).

For the prediction-based algorithm, the same phenomenon can still be observed, but it is much less pronounced. Lane usage is distributed more evenly across the network, and the under-utilization of lane (2,0) is largely mitigated.

Since the position cost function encourages drones whose destination is not the off-ramp to move away from lane (2,0), these drones tend to concentrate in the left-hand-side lanes under both algorithms. As lanes such as (0,0) and (-2,0) become increasingly crowded, drones using STDG are less likely to switch to another lane because they assume that many neighboring drones have the same intention. In contrast, the prediction-based algorithm enables drones to predict the actions of their neighbors, and allows them to avoid such overly cautious decision-making and make effective use of the available road space.

In the 1r1r scenario, the average speeds achieved by the two algorithms are both around 5 m/s, indicating that the severe congestion limits the effectiveness of the prediction-based algorithm. In contrast, in the 1r0r scenario, the average speeds achieved by the STDG and prediction-based algorithms are 11.12 m/s and 12.48 m/s, respectively. Although the prediction-based algorithm provides only a modest improvement over STDG, it maintains an average speed that is very close to the ideal cruising speed of 12.5 m/s.

\subsection{Complex Road System and Impact of Block Length}

\begin{figure}[htbp]
  \centering
  \includegraphics[scale=0.6]{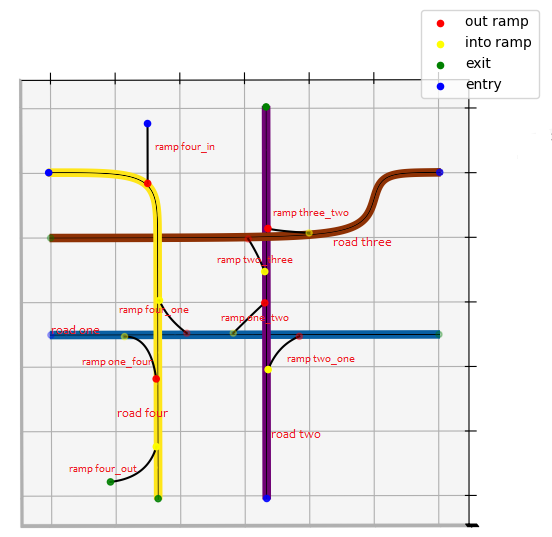}
  \caption{Drone Road System used in the simulations}
  \label{fig:drsInSim}
\end{figure}

In the following studies, we focus on evaluating the prediction-based algorithm on a complex road system which contains four roads, seven ramps, and a total of 21 lanes. The road system is shown in Figure \ref{fig:drsInSim}. Each unit cell in the figure represents a 200 m × 200 m area, and all four roads are about 1200 meters long.

\begin{figure}[htb!]
  \centering
  \includegraphics[scale=0.3]{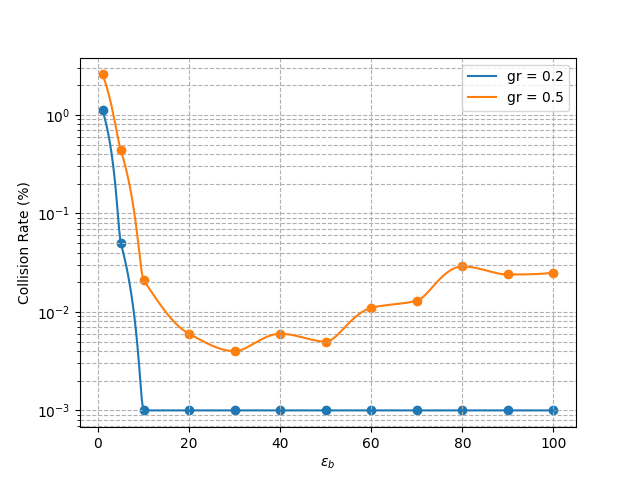}
  \caption{Collision rate for different block lengths}
  \label{fig:ebcr}
\end{figure}

\begin{figure}[htb!]
  \centering
  \includegraphics[scale=0.3]{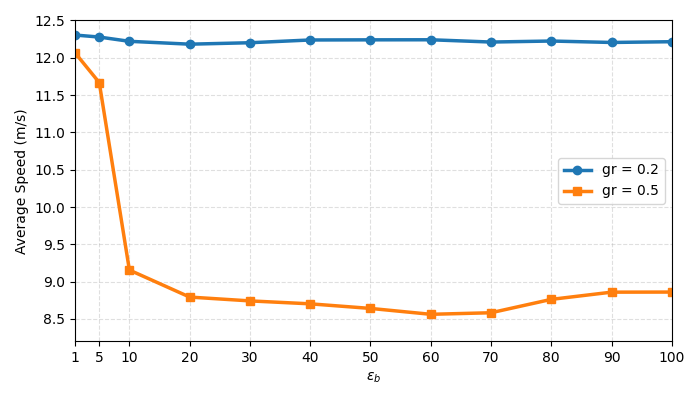}
  \caption{Average speed for different block lengths}
  \label{fig:ebas}
\end{figure}

We investigate the effect of the block length $\epsilon_b$ on the performance metrics, where $\epsilon_b$ is measured in meters. The parameter combination $\kappa_1$-$\kappa_2$-$\epsilon_1$ = 10-10000-15 is used throughout this experiment, and two drone generation rates, gr=0.2 and gr=0.5, are considered. The results are presented in Figures \ref{fig:ebcr} and \ref{fig:ebas}. For the $gr=0.2$ case, the collision rate reaches its highest value 1.11\% when $\epsilon_b=1m$, and rapidly decreases to a close-to-zero level as $\epsilon_b$ increases. The average speed shows no significant variation, remaining around 12.26 m/s for all tested values. 

For the $gr=0.5$ case, the collision rate remains significantly high when $\epsilon_b$ is small. The highest collision rate of 2.58\% occurs when $\epsilon_b=1m$, while the second-highest value of 0.44\% is observed when $\epsilon_b=5m$. However, when $\epsilon_b > 50m$, only a few collisions are observed, and the difference between these cases is negligible. Regarding the average speed, it decreases rapidly from 12.06 m/s to 8.79 m/s when $\epsilon_b < 30m$, and then remains relatively stable as $\epsilon_b$ continues to increase.

\subsection{Further Comparison and Sensitivity Analysis}
\label{sec:results:sensitivity}

The next simulation compares the proposed prediction-based algorithm with STDG. Based on the previous results for the STDG algorithm, when $gr = 0.2$, the best collision rate performance is achieved with the $\kappa_1$-$\kappa_2$-$\epsilon_1$ combination of 1-10000-15, while the highest average speed is achieved with the 100-1-5 combination. We then investigate the performance of these two parameter combinations under the prediction-based algorithm. Table \ref{tab:complex result} summarizes the results for both algorithms.

\begin{table}[ht]
\centering
\footnotesize
\begin{tabular}{|l|l|l|l|l|l|l|}
\hline
& \textbf{$\kappa_1$} & \textbf{$\kappa_2$} & \textbf{$\epsilon_1$} & \textbf{Max.  Collision Rate} & \textbf{Avg. Speed (m/s)} & \textbf{Arrival Count} \\ \hline
\textbf{STDG}             & \multirow{2}{*}{1}                & \multirow{2}{*}{10000}            & \multirow{2}{*}{15}                &       0.057\%       &       10.2842     &      2.74         \\ \cline{1-1} \cline{5-7} 
\textbf{Prediction-Based} &                                   &                                   &                                      &      0.00\%       &    12.1829          & 3.62       \\ \hline
\textbf{STDG}             & \multirow{2}{*}{100}              & \multirow{2}{*}{1}                & \multirow{2}{*}{5}                   &       1.27\%      &   12.3005           &   3.73          \\ \cline{1-1} \cline{5-7} 
\textbf{Prediction-Based} &                                   &                                   &                                      &    0.64\%         &     12.3117       &       3.73      \\ \hline
\textbf{Ideal} &                                   &                                   &                                      &    0.00\%         &     12.5       &       4.2      \\ \hline
\end{tabular}
\caption{Simulation results for 50 replications in the complex road system for STDG and PB-STDG, where gr = 0.2}
\label{tab:complex result}
\end{table}

In the first case ($\kappa_1 = 1$, $\kappa_2 = 10000$, $\epsilon_1 = 15$), which corresponds to a low-collision configuration, both algorithms achieve a close-to-zero collision rate. However, the prediction-based algorithm achieves a noticeably higher average speed (12.18 m/s compared to 10.28 m/s for STDG) and approaches the theoretical ideal value of 12.5 m/s. It also achieves a higher arrival count, indicating improved overall system throughput without increasing collision risk.

The arrival count represents the system throughput, and the ideal value under the tested setting is 4.2. The prediction-based algorithm achieves 3.62 arrivals, compared to 2.74 for STDG, corresponding to 86.2\% and 65.2\% of the ideal value, respectively.

In the second case ($\kappa_1 = 100$, $\kappa_2 = 1$, $\epsilon_1 = 5$), which emphasizes speed-oriented optimization, both algorithms reach near-maximum average speeds (around 12.30 m/s). But the prediction-based algorithm still achieves a lower maximum collision rate (0.64\% versus 1.27\%). Compared to the ideal metrics, the parameter combination (1, 10000, 15) for the prediction-based algorithm provides a favorable balance between safety and traffic efficiency among the tested settings, achieving a zero collision rate and an average speed of 12.2 m/s, only 0.3 m/s below the theoretical ideal value.

We further test more parameter combinations and conduct a sensitivity study on the prediction-based algorithm to understand the relative impact of these parameters on our performance metrics. For comparison, the same parameter combinations are also evaluated using STDG. Table \ref{tab:complex result full} shows the full simulation results under various parameter combinations, and Table \ref{tab:regression_result} shows the results of the sensitivity study.

\begin{table}[ht]
\centering
\footnotesize
\begin{tabular}{|r|r|r|r|r|r|r|}
\hline
\multicolumn{1}{|l|}{\textbf{Generation Rate}} &
\multicolumn{1}{|l|}{\textbf{$\kappa_1$}} &
  \multicolumn{1}{l|}{\textbf{$\kappa_2$}} &
  \multicolumn{1}{l|}{\textbf{$\epsilon_1$}} &
  \multicolumn{1}{l|}{\textbf{Collision Rate  (\%)}} &
  \multicolumn{1}{l|}{\textbf{Average Speed (m/s)}} &
  \multicolumn{1}{l|}{\textbf{Arrival Count}} \\ \hline
0.2 & 1  & 10    & 15 & 0.0086 (1.3888) & 12.32 (9.56)& 3.73 (1.69)\\ \hline
0.2 & 1  & 10    & 5  & 0.4224 (1.41) & 12.36 (12.01) & 3.78 (3.79) \\ \hline
0.2 & 1  & 10000 & 15 & 0 (0.0026)  & 12.18 (10.28) & 3.62 (2.74) \\ \hline
0.2 & 1  & 10000 & 5  & 0.4694 (1.164)& 12.37 (11.72)& 3.75 (3.57)\\ \hline
0.2 & 10 & 10    & 15 & 0.0022 (1.1604)& 12.28 (11.89)& 3.74 (3.64)\\ \hline
0.2 & 10 & 10    & 5  & 0.314  (1.726)& 12.28 (11.93)& 3.78 (3.57)\\ \hline
0.2 & 10 & 10000 & 15 & 0       (0.0048) & 12.28 (11.13)& 3.75 (3.58)\\ \hline
0.2 & 10 & 10000 & 5  & 0.3306 (1.0864)& 12.29 (11.83) & 3.78 (3.41)\\ \hline
0.1 & 1  & 10    & 15 & 0      (0)  & 12.47 (12.17)& 2.01 (1.99)\\ \hline
0.1 & 1  & 10    & 5  & 0.138 (0.1946)& 12.48 (12.24)& 1.97 (1.95)\\ \hline
0.1 & 1  & 10000 & 15 & 0      (0)  & 12.49 (11.95)& 1.98 (1.98)\\ \hline
0.1 & 1  & 10000 & 5  & 0.0934 (0.3966)& 12.49 (12.14)& 1.98 (1.91)\\ \hline
0.1 & 10 & 10    & 15 & 0.004  (0.3680)& 12.45 (12.24)& 1.99 (1.99)\\ \hline
0.1 & 10 & 10    & 5  & 0.0646 (1.4428)& 12.41 (12.30)& 1.96 (1.98)\\ \hline
0.1 & 10 & 10000 & 15 & 0      (0)  & 12.46 (11.95)& 1.99 (1.97)\\ \hline
0.1 & 10 & 10000 & 5  & 0.1003 (0.4016)& 12.43 (12.11)& 2  (2.01)  \\ \hline
\end{tabular}
\caption{Simulation results from 50 independent simulation replications under various parameter combinations for PB-STDG and STDG (shown in parentheses).}
\label{tab:complex result full}
\end{table}

Across different generation rates, the overall average speed and arrival count for these parameter combinations remain relatively consistent. When $gr = 0.2$, the average speed remains around 12.3 m/s, while for $gr = 0.1$, it is approximately 12.45 m/s.

The collision rate becomes zero in many cases where $\epsilon_1 = 15$ m, whereas in the remaining cases it remains close to zero ( $<9\times10^{-3}\%$). In contrast, when $\epsilon_1 = 5$ m, the collision rate shows noticeable non-zero values ranging from $0.06\%$ to $0.5\%$.

Compared with the STDG algorithm, the prediction-based algorithm consistently achieves a higher average speed while maintaining a lower or comparable collision rate across all parameter combinations. In particular, in the 1-10000-15 and 10-10000-15 scenarios, both algorithms achieve close-to-zero collisions, whereas the prediction-based algorithm provides a noticeable improvement in both average speed and throughput.

\begin{table}[ht]
\centering
\footnotesize
\setlength{\tabcolsep}{3pt}
\renewcommand{\arraystretch}{0.9}
\begin{tabular}{|l|l|l|l|}
   \hline 
   & \textbf{Collision rate}  & \textbf{Average speed (m/s)} & \textbf{Arrival count} \\ 
   \hline 
   Average of responses & 1.2e-3 & 12.3837 & 2.8631 \\ 
   \hline 
   Minimum of responses & 0.0 & 12.1829 & 1.97 \\ 
   \hline 
   Maximum of responses & 4.69e-3 & 12.4952 & 3.81 \\ 
   \hline 
   $R^2$ value (\%) & 99.1394 & 94.5068 & 99.9791 \\ 
   \hline 
   \multicolumn{4}{|l|}{\textbf{Regression coefficients}} \\ 
   \hline 
   Cost parameter $\kappa_1$ (Coefficient $\alpha_1$) & -2.0e-4 & -1.8e-2 & 1.1e-2 \\ 
   \hline 
   Cost parameter $\kappa_2$ (Coefficient $\alpha_2$) & 0.0 & -3.8e-3 & -6.9e-3 \\ 
   \hline
   Distance threshold parameter $\epsilon_1$ (Coefficient $\alpha_3$) & -1.2e-3 & -1.0e-2 & -1.2e-2 \\ 
   \hline  
   Generation rate (Coefficient $\alpha_4$) & -8.0e-4 & -8.4e-2 & 0.8781 \\ 
   \hline 
   \multicolumn{4}{|l|}{\textbf{Percentage contributions}} \\ 
   \hline 
   Cost parameter $\kappa_1$ (\% contr.) & 1.2934 & 3.5112 & 0.0146 \\
   \hline 
   Cost parameter $\kappa_2$ (\% contr.) & 0.0626 & 0.1567 & 0.0061 \\
   \hline 
   Distance threshold parameter $\epsilon_1$ (\% contr.) & 54.4785 & 1.1312 & 0.0183 \\ 
   \hline 
   Generation rate (\% contr.) & 20.8291 & 78.4950 & 99.8278 \\ 
   \hline 
\end{tabular}
\caption{Regression analysis results and percentage contributions of parameters for PB-STDG.}
\label{tab:regression_result}
\end{table}

A linear regression model was fitted using coded parameter levels to quantify the relative influence of the tested factors. The remaining variation of percentage contribution is associated with residual and/or interaction effects.

From the sensitivity study results of the PB-STDG, the $R^2$ values for collision rate, average speed, and arrival count are 99.14\%, 94.51\%, and 99.98\%, which indicates that the regression models fit the simulation data very well and are able to explain most of the observed variation in the performance metrics.

The generation rate contributes most significantly to the arrival count (99.83\%). It also has a substantial effect on the average speed (78.50\%) and a moderate effect on the collision rate (20.83\%). Among the algorithm parameters, the distance threshold parameter $\epsilon_1$ makes the largest contribution to the collision rate (54.48\%), with a negative regression coefficient indicating that increasing $\epsilon_1$ reduces the collision rate. In addition, $\epsilon_1$ has a minor negative effect on the average speed (1.13\%).

The remaining regression coefficients and percentage contributions indicate that the position cost parameter $\kappa_1$ and the collision cost parameter $\kappa_2$ have limited influence on the performance metrics. Specifically, $\kappa_1$ contributes 3.51\% to the average speed and 1.29\% to the collision rate, while the contributions of the other parameters are negligible.

%\begin{tabular}{|l|l|l|l|}
%\hline
% \textbf{Regression coefficients}   & \textbf{Collision rate}  & \textbf{Average speed (m/s)} & \textbf{Arrival count}\\ 
%hline
% $\kappa_1\times \kappa_2$ &  0.0e-3 &  0.0069 &  0.0131  \\ 
%\hline
% $\kappa_1\times \epsilon_1$ &  0.1e-3 &  0.0168 &  0.0056  \\ 
%\hline
% $\kappa_1\times gr$ &  -0.1e-3 &  0.0062 &  0.0106  \\ 
%\hline
% $\kappa_2\times \epsilon_1$ &  -0.0e-3 &  -0.0097 &  -0.0094  \\ 
%\hline
% $\kappa_2\times gr$ &  0.0e-3 &  -0.012 &  -0.0094  \\ 
%\hline
% $\epsilon_1\times gr$ &  -0.4e-3 &  -0.02 &  -0.0194  \\ 
%\hline
 %\textbf{Percentage contributions}   & \textbf{Collision rate}  & %\textbf{Average speed (m/s)} & \textbf{Arrival count}\\ 
%\hline
 %$\kappa_1\times \kappa_2$ (\% contr.) &  0.0121 &  0.5344 &  0.0223  \\ 
%\hline
 %$\kappa_1\times \epsilon_1$ (\% contr.) &  1.2512 &  3.1454 &  0.0041  \\ 
%\hline
% $\kappa_1\times gr$ (\% contr.) &  0.7048 &  0.4284 &  0.0146  \\ 
%\hline
 %$\kappa_2\times \epsilon_1$ (\% contr.) &  0.1336 &  1.0486 &  0.0114  \\ 
%\hline
% $\kappa_2\times gr$ (\% contr.) &  0.026 &  1.5981 &  0.0114  \\ 
%\hline
 %$\epsilon_1\times gr$ (\% contr.) &  20.3481 &  4.4578 &  0.0486 % \\ 
%\hline
%\end{tabular}

\subsection{Discussion}
\label{sec:results:discussion}

In the simulation study, we investigate both the previous STDG algorithm and the proposed prediction-based algorithm under various parameter settings across several road systems. Across all simulation results, a common conclusion is that the prediction-based algorithm consistently achieves higher average speeds while maintaining comparable or lower collision rates compared to the STDG algorithm in all tested scenarios.

This advantage becomes more pronounced when the drone density is high or when a large proportion of nearby drones need to switch lanes frequently. For example, in the complex drone road system scenario, the average speed increases from 10.28 m/s to 12.18 m/s when using the prediction-based algorithm. In the scenario shown in Figure \ref{fig:2r2r_as}, the prediction-based algorithm provides at least a 35\% improvement in average speed without degrading the collision rate.

The additional prediction mechanism provides a possible explanation for these observations. During the previous study, the challenge of STDG in the feasible-lane selection phase was that drones could not determine whether nearby drones also intended to switch to the same lane. Under the low-collision-rate strategy (i.e., low position cost weight and high collision cost weight), drones tend to make highly conservative decisions. A candidate lane can be easily discarded when nearby drones are detected in its neighboring lanes, causing the stay-in-current-lane behavior to become the preferred option. This consequently reduces the overall average speed. Under the speed-oriented strategy (i.e., high position-cost weight and low collision-cost weight), the issue described above can be alleviated. However, the trade-off is that drones tend to ignore the risk of concurrent switching, which consequently leads to a higher collision risk.

PB-STDG is able to balance these two strategies. Since drones share additional information with neighboring drones, such as target lanes and preferred speeds, they can predict the decisions of nearby drones and make more informed decisions accordingly. In the evaluated scenario, only a small number of drones near the front boundary of a block need to explicitly consider concurrent switching conflicts. Most drones instead rely on predicted behaviors of neighboring drones and select either the optimal lane or a secondary optimal lane if the preferred lane is already occupied. Drones then have a greater opportunity to switch to lanes that either allow higher cruising speeds or lead toward their target destinations without unnecessary slowing or stopping. To further validate this finding, we collect the total number of algorithm executions and the number of decisions that switch lanes for both algorithms in the same four-road, seven-ramp DRS with the same parameter configuration (10-10000-15 as $\kappa_1$-$\kappa_2$-$\epsilon_1$, and $gr=0.2$). The results are shown in Table \ref{tab:concurrent switching}.

\begin{table}[ht!]
  \centering
  \renewcommand{\arraystretch}{0.8}
  \begin{tabular}{|l|l|l|}
    \hline
    \textbf{Algorithm}              & \textbf{Total Algorithm Executions}    & \textbf{Lane-Switching Decisions}                     \\ \hline
    STDG        &  186288  & 2445                 \\ \hline
    PB-STDG         &  142419  &  4790            \\ \hline
  \end{tabular}
  \caption{Comparison of STDG and PB-STDG in Handling Concurrent Lane Switching }
  \label{tab:concurrent switching}
\end{table}

From the results, the prediction-based algorithm performs fewer algorithm executions but makes more lane-switching decisions. According to the cost model, a successful lane-switching action either increases a drone's cruising speed or moves it closer to its target lane. Both cases help the drone reach its destination faster and consequently improve the throughput. Therefore, the substantial increase in lane-switching decisions made by the prediction-based algorithm indicates that it can encourage drones to safely perform lane switching, thereby improving traffic efficiency.

In addition, since the number of drones is similar for both algorithms and drones execute the algorithm more frequently when they are in the stop state, fewer algorithm executions suggest that drones using the prediction-based algorithm are less likely to remain in the stop state. This indirectly demonstrates the higher operational efficiency of drones under the prediction-based algorithm.

Based on the additional simulation results on lane usage distribution, the previously proposed inner lane vanishing hypothesis is not an accurate explanation of the limitation of the STDG algorithm. The results suggest that, without any lane-switching demand, the STDG algorithm does not noticeably affect the lane usage distribution and therefore does not cause inner lanes to vanish. However, when a large proportion of drones share the same target lane, the STDG algorithm can cause the target lane to gradually fall out of use. This occurs because many drones fail to switch to the desired lane in order to avoid potential collisions with other drones attempting the same maneuver. As a result, these drones may accumulate in the surrounding lanes and block the target lane.

Figure \ref{fig:ld_1r1r} indicates that the PB-STDG algorithm can mitigate the lane vanishing problem by applying the same mechanism for resolving concurrent switching conflicts discussed above. In addition, it leads the lane usage distribution toward a more even pattern, but high-demand lanes still have correspondingly high usage. In contrast, under the STDG algorithm, some high-demand lanes remain underutilized. This is one of the main reasons why the prediction-based algorithm achieves a higher average speed.

However, in an extremely crowded scenario, where a large number of drones have the same target ramp and become congested before entering it, the prediction-based algorithm provides limited improvement in the throughput from the road to the target ramp. In this case, the bottleneck is no longer caused by the decision-making process but by the limited capacity of the drone road system, which is determined by drone density and road system design.

Based on the results of varying the block length $\epsilon_b$, $\epsilon_b$ should not be set to a small value, especially when it is smaller than $\epsilon_1$. This is because drones only consider neighbors belonging to the same or adjacent blocks as potential collision risks. When $\epsilon_b$ is small, drones only consider a relatively limited area (their current and adjacent blocks) around themselves and may ignore other drones that are physically close but located outside the considered blocks. In this case, the collision rate increases noticeably. When the drone density is high, a larger $\epsilon_b$ value may slightly increase the collision rate. This is because a larger block size means that more neighboring drones need to be considered during the prediction process. An incorrect prediction for any neighboring drone may affect the decision-making process and eventually lead to a collision. Therefore, an appropriate value of $\epsilon_b$ should be larger than the threshold $\epsilon_1$ but should not be excessively large, as it may increase the prediction complexity and prediction error rate.

The sensitivity study indicates that the environmental parameter generation rate has the greatest influence on average speed and throughput, while the algorithm parameter $\epsilon_1$ has the most significant effect on collision rate. In comparison, the remaining parameters, $\kappa_1$ and $\kappa_2$, have only limited influence on the performance metrics in the prediction-based algorithm. Under the STDG algorithm, however, $\kappa_1$ and $\kappa_2$ contribute more substantially to system performance.

This phenomenon is expected for both algorithms. In the STDG algorithm, the $\kappa_1/\kappa_2$ ratio controls the strategy used to handle concurrent switching problems. A higher ratio leads to more aggressive decisions, whereas a relatively lower ratio results in more conservative behavior. These different strategies then produce variations in both collision rate and average speed. In the prediction-based algorithm, $\kappa_2$ is often less influential because the movements of nearby drones can be predicted, making it unnecessary to heavily consider potential collision risks in most cases. As a result, varying $\kappa_1$ alone does not substantially alter the candidate lane pool. Consequently, the average speeds across different parameter settings remain close to each other.

Although the prediction-based algorithm performs better than the STDG algorithm in our evaluated scenarios, it still has several disadvantages that are not directly reflected in the simulation results. The first issue is the increased computational complexity. For STDG, the computational complexity depends on the number of nearby drones and is affected by the transmit power. Under the same transmit power, since drones are required to predict the movements of nearby drones within the same block, the overall complexity increases from $O(n)$ to $O(mn)$, where $n$ is the number of nearby drones and $m$ is the number of nearby drones within the same block. In addition, a relatively minor issue is the vulnerability introduced by the additional information shared in beacon packets. Such information may be manipulated to gain higher lane switching priority. For example, a drone could falsely advertise a higher preferred speed, thereby increasing the speed cost calculated by neighboring drones and causing them to avoid a lane that the malicious drone subsequently switches into.

\section{Conclusion and Future Work}
\label{sec:conclusion}

This paper presents PB-STDG, a prediction-based extension of the Short-Term Decentralized Greedy (STDG) guidance algorithm. The proposed algorithm inherits the cost-function mechanism of STDG, which encourages drones to continuously make short-term greedy decisions to improve traffic efficiency. It further introduces a prediction mechanism that enables drones to anticipate the decisions of neighboring drones using the additional information shared in beacon packets. This prediction process alleviates the over-conservative behavior observed in high-density traffic scenarios. By predicting the intended movements of neighboring drones, each drone can make more informed decisions. 

A series of simulation studies was conducted to evaluate the proposed prediction-based algorithm and compare its performance with the original STDG algorithm under the same drone road system. We first investigated the effects of wireless communication parameters, including the beaconing rate and transmit power. The results show that, within a reasonable range, these parameters have little impact on the algorithm performance. We then compared the two algorithms under different drone road system configurations and algorithm parameter settings. The simulation results demonstrate that the proposed PB-STDG consistently achieves higher average speed and throughput while maintaining the same or even lower collision rate in almost all scenarios.

The sensitivity analysis and the block length study further show that the influence of the original cost-function weights, $\kappa_1$ and $\kappa_2$, is noticeably reduced in the prediction-based algorithm. Instead, the performance is mainly determined by the distance threshold $\epsilon_1$ and the newly introduced block length parameter $\epsilon_b$. With appropriate parameter settings, the algorithm achieves an average speed close to the ideal cruising speed while maintaining a close-to-zero collision rate.

The main concern for the prediction-based algorithm is its computational complexity, which depends on the number of drones within the same block. Choosing a smaller value of $\epsilon_b$ can reduce the additional computational cost, but it may also increase the collision rate because the prediction process may fail to include all nearby drones. Another potential issue is the vulnerability introduced by the additional shared information. Malicious drones may transmit false information to gain priority during lane-switching competition. Finally, this study assumes ideal operating conditions by neglecting GPS positioning errors, interference from other wireless signals, environmental noise, and ionospheric or tropospheric disturbances. Therefore, the simulation results may not fully represent real-world scenarios.

Future research may address these limitations in several directions. One possible approach is to develop a purely coordination-based or negotiation-based guidance algorithm that avoids complex prediction and cost calculations. Alternatively, the prediction mechanism could be further optimized to reduce computational complexity while incorporating data validation or trust mechanisms to improve the reliability of the shared information. In addition, more realistic communication and positioning models should be incorporated into the simulation to evaluate the robustness and applicability of the proposed algorithm under practical operating conditions. To further improve the realism of the simulation, developing more accurate models of drone dynamics and the lane-switching process is another possible direction.

\section*{Acknowledgment}

This project is supported by University of Canterbury Aho Hinatore
Accelerator Scholarship. X. Wu is partially supported by the Tait Foundation. 

\section*{AI Declaration}

Generative artificial intelligence (AI) tools were used only for language editing and grammar correction. They were not used to generate scientific content, research ideas, methodology, results, or conclusions. All AI-assisted revisions were carefully reviewed and verified by the authors, who take full responsibility for the final content of the manuscript.

% =======================================================================
\bibliography{paper}
% =======================================================================
\end{document}